\documentclass[12pt]{article}

\usepackage{fontspec}
\usepackage{xeCJK}

\setCJKsansfont[
  BoldFont={FandolHei-Bold.otf}
]{FandolHei-Regular.otf}

\setCJKmonofont{FandolFang-Regular.otf}

\usepackage[margin=1in]{geometry}
\usepackage{graphicx}
\usepackage{subcaption}
\usepackage{float}
\usepackage{placeins}
\usepackage{xcolor}
\usepackage{tabularx}
\usepackage{array}
\usepackage{amsmath}
\usepackage{bm}
\usepackage{booktabs}
\usepackage{lineno}
\usepackage[hidelinks]{hyperref}

\usepackage[
  backend=biber,
  bibstyle=biblatex-sp-unified,
  citestyle=sp-authoryear-comp,
  maxcitenames=3,
  maxbibnames=99,
  natbib
]{biblatex}

\hypersetup{
  pdfauthor={},
  pdftitle={Reduplicative constructions in Mandarin:
  Socio-emotional profiling through distributional semantics}
}

\title{Reduplicative constructions in Mandarin:
Socio-emotional profiling through distributional semantics}

\author{%
  Chaoyi Wu\textsuperscript{1},
  Yu-Hsiang Tseng\textsuperscript{2},
  R. Harald Baayen\textsuperscript{2}\\[1em]
  \small \textsuperscript{1}Beihang University\\
  \small \textsuperscript{2}University of Tübingen
}
\date{}

\begin{document}

\maketitle

\begin{abstract}
\noindent
Mandarin Chinese has two productive reduplicative constructions that repeat either two-character base words or their constituents (e.g., 健健康康 `in good health', 讨论讨论 `discuss a bit'). Their varied meanings have been described as realizing plurality, valence coloring, sound symbolism and pragmatic functions. The aim of this study is twofold.  A first goal is to clarify whether it is possible to come to a more precise understanding of the variegated semantics of Mandarin reduplication by using word embeddings from distributional semantics.  A second goal is to explore how useful embeddings are for understanding the details of a semantically complex word-formation process. We show that the embedding space recovers the semantic and grammatical properties of reduplications previously identified in the literature, validating Tencent embeddings for morphological investigation. Semantic profiling revealed that reduplicative constructions are often strongly represented on multiple dimensions. The two patterns exhibit clear semantic and pragmatic differentiation in distributional space. Procrustes analysis clarified that the overall organization of the base-word space is largely preserved in the reduplication space, with local mismatches highlighting regions of discourse-pragmatic reorganization. Taken together, these results show that high-dimensional word embeddings can recover established linguistic generalizations, and capture the semantic versatility of Mandarin reduplication and constructional transparency.
\end{abstract}


\section{Introduction}

This study presents a quantitative investigation of two four-syllable reduplicative constructional patterns in Mandarin Chinese  which are derived from disyllabic bases. Using word embeddings, we investigate how these reduplicative constructions are distributed in semantic space, what semantic changes distinguish them from their base words, and to what extent the reduplicative constructions remain semantically transparent with respect to their bases. Our study builds on previous research in Mandarin linguistics on reduplication \citep{chao1968spoken,LiThompson1981,ZhuDexi1982, hua2003reduplication,Wang2023,lu2026deliminative}, as well as on previous quantitative work on word and construction formation using distributional semantics \citep{MarelliBaroni2015,PerekHilpert2017,YangBaayen2026}, research on semantic shifts in inflectional and derivational morphology \citep{NikolaevEtAl2022,StupakBaayen2022, ShafaeiBajestanEtAl2024}, studies of semantic transparency in Mandarin lexicons \citep{ShenBaayen2022a,ShenBaayen2022b}, and work comparing lexical semantic spaces using Procrustes analysis \citep{YangBaayen2025}.

The two reduplicative constructions investigated in the present study repeat either the individual constituents (henceforth the AABB construction) or the whole word (henceforth the  ABAB construction).  For example, the base 健康 \textit{jian4-kang1} `health, healthy' gives rise to the AABB construction 健健康康 \textit{jian4-jian4-kang1-kang1} `in good health'.  For 讨论讨论 (\textit{tao3-lun4-tao3-lun4} `discuss a bit'), by contrast, the disyllabic base 讨论 (\textit{tao3-lun4} `discuss') is reduplicated as a whole. 

In previous standard grammatical descriptions, the reduplicative constructions have been associated with a bewilderingly wide range of semantic and grammatical properties. Compared with their disyllabic bases, reduplicative constructions have been described as expressing meanings such as plurality, degree intensification, affective coloring, or the pragmatics of persuasive force \citep{chao1968spoken, hua2003reduplication}. Furthermore, the AABB and ABAB patterns show different tendencies: AABB reduplications have been associated with plurality of entities in nominal uses, iteration of events in verbal uses \citep{Zhang2015}, and intensification of degree in adjectival uses \citep{ZhuJingsong2003, MelloniBasciano2018}. ABAB constructions, on the other hand, especially in verbal uses, are commonly associated with tentative or suggestive interpretations \citep{lu2026deliminative}.

In the present study, we investigate to what extent the semantic and pragmatic properties attributed to Mandarin reduplication can be recovered in distributional semantic space. In other words, we ask whether high-dimensional word embeddings capture linguistically interpretable aspects of the meanings of Mandarin reduplicative constructions, whether AABB and ABAB show different distributions in semantic space, what changes in meaning arise when going from a disyllabic lexical base to its derived reduplicative forms, and to what extent the reduplicative constructions remain semantically transparent with respect to their base words, by comparing the semantic spaces of the base-word and reduplicative constructions.

To address these questions, we extract all the Mandarin reduplicative constructions from the Center for Chinese Linguistics Corpus \citep[CCL Corpus;][]{zhan2019ccl}, represent their meanings and those of their disyllabic bases using 200-dimensional Tencent embeddings \citep{song2018directional}, and use the Mandarin non-reduplicative words investigated by \citet{YangBaayen2025} as a reference baseline. We first examine how reduplicative constructions are distributed in semantic space and profile these constructions in terms of the linguistically semantic and grammatical properties. We then represent the change from a base word to its derived reduplicative construction as a shift vector and investigate what these semantic shifts are. Finally, we compare the base-word and reduplication spaces using Procrustes analysis in order to assess their global correspondence and identify local departures from semantic transparency.

In the remainder of this study, we address the following research questions:

\begin{enumerate}
    \item Do Mandarin reduplicative constructions form a distinct and  internally structured domain in the broader Mandarin semantic space?

    \item How are reduplicative constructions profiled along linguistically grammatical and semantic properties, and to what extent do the AABB and ABAB patterns differ in terms of these properties?

    \item What semantic changes arise when going from a disyllabic base word to its derived reduplicative construction?

    \item To what extent are reduplicative constructions semantically transparent with respect to their base words?
\end{enumerate}

The remainder of this paper is structured as follows. Section~\ref{sec:data} introduces the dataset and the semantic representations used in the study. Section~\ref{sec:Clustering} examines the clustering structure of reduplicative constructions in semantic space. Section~\ref{sec:semantic-profiling} characterizes these clusters further in terms of semantic profiling, including lexical category, valence, self-relatedness, and sound-related meaning. Section~\ref{sec:shift-vectors} investigates the semantic shifts between reduplicative constructions and their identifiable base words. Section~\ref{sec:procrustes} compares the semantic structures of the base-word space and the reduplication space by means of Procrustes analysis. Section~\ref{sec:discussion} concludes with a general discussion.

\section{Data}
\label{sec:data}

From the corpus compiled by the Center for Chinese Linguistics (CCL; \citealt{zhan2019ccl}), we extracted all four-character reduplications with the AABB and ABAB patterns.  Of the 1,010 reduplications,  520 instantiated the AABB pattern and 490 the ABAB pattern. We collected the corresponding embeddings for these reduplications from the Tencent repository \citep{song2018directional}. These embeddings make use of the algorithms underlying word2vec \citep{mikolov2013distributed}.  

We then identified the corresponding base word for each reduplication. In most cases, this base word is the disyllabic form AB underlying the reduplicative construction. For example, the base word of 健健康康 \textit{jian4-jian4-kang1-kang1} `in good health' is 健康 \textit{jian4-kang1} `healthy', and the base word of 讨论讨论 \textit{tao3-lun4-tao3-lun4} `discuss a bit' is 讨论 \textit{tao3-lun4} `discuss'.

However, not every reduplication has a corresponding base word. For instance, 盆盆罐罐 \textit{pen2-pen2-guan4-guan4}, `pots and pans', refers to kitchen utensils, but the disyllabic form 盆罐 \textit{pen2-guan4} is not an existing word. For such cases, no base word embedding is available.  Some reduplications have base words that are linguistically implausible. For example, for 林林总总 \textit{lin2-lin2-zong3-zong3}, meaning `various and miscellaneous', 林总 \textit{lin2-zong3} is attested, but its meaning is unrelated: `Manager Lin'. These forms were not included in the set of base words. The resulting dataset contained 955 reduplication-base pairs involving 916 distinct base words.

It should be mentioned that some reduplications share the same base word while differing in reduplicative construction. For example, the base word 热闹 \textit{re4-nao} `lively' corresponds to both the AABB form 热热闹闹 \textit{re4-re4-nao-nao} `very lively' and the ABAB form 热闹热闹 \textit{re4-nao-re4-nao} `liven up'. In total, 39 base words in the dataset give rise to both possible constructions.

\section{Clusters of reduplicative constructions}
\label{sec:Clustering}

Before examining the semantic distribution of reduplicative constructions, we first asked whether they occupy identifiable regions in a broader Mandarin lexical semantic space. To this end, we combined the 1,010 reduplicative constructions in our dataset with the Mandarin reference word set analyzed by \citet{YangBaayen2025}, which contains 2,173 non-reduplicated Mandarin words from 21 semantic categories.

\begin{figure}[H]
\centering
\includegraphics[width=0.5\linewidth]{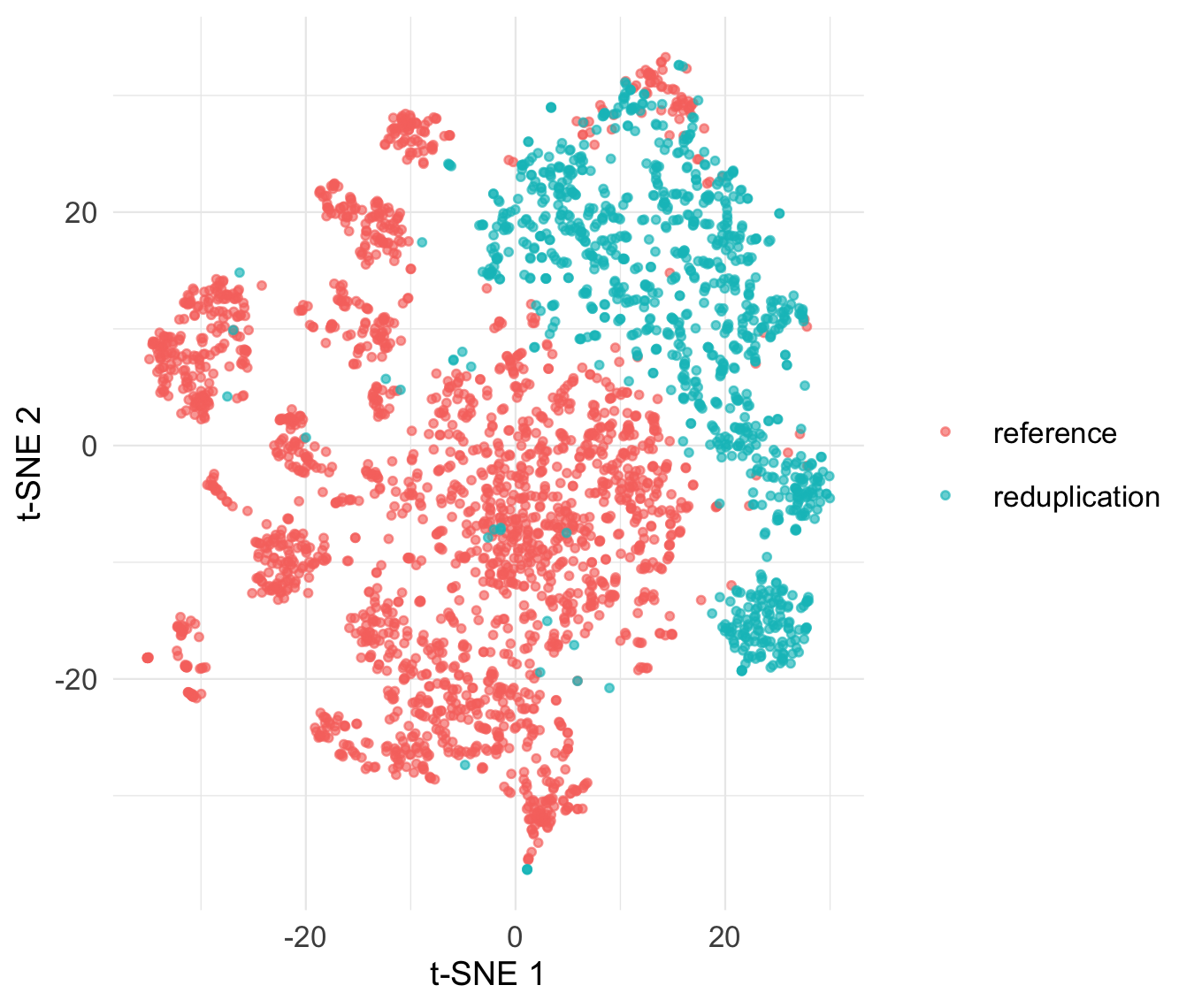}
\caption{Two-dimensional t-SNE plot situating reduplications in a broader lexical semantic space together with Mandarin reference words. Reduplications occupy several identifiable regions rather than being randomly scattered among the reference words.}
\label{redup_in_broader_space}
\end{figure}

As shown in Figure~\ref{redup_in_broader_space}, reduplicative constructions are not randomly scattered among the reference words, but occupy several identifiable regions of the t-SNE plane. To further assess this separation, we used linear discriminant analysis to test whether the embeddings distinguish reduplicative constructions from the reference words. Under leave-one-out cross-validation, the LDA achieved an accuracy of 98.7\%, substantially above the majority-class baseline of 68.3\%. These results prove that reduplicative constructions are distributionally distinguishable from non-reduplicated Mandarin words in the broader lexical semantic space.

To address our first research question, which asks whether reduplication constructions form distinct clusters in distributional semantic space, we applied k-means clustering \citep{macqueen1967some} to the 200-dimensional Tencent embeddings of all reduplications in our dataset. This allowed us to examine whether semantically similar Mandarin reduplications tend to cluster together rather than being randomly dispersed across the vector space. We evaluated several values of $k$ and selected $k=10$ as the final clustering solution. This choice was supported by a subsequent linear discriminant analysis \citep{VenablesRipley2002} with leave-one-out cross-validation, which achieved an accuracy of 90.50\%, see Table~\ref{tab:LDA} (majority baseline: 16.24\%). Table~\ref{tab:cluster_examples} presents two characteristic examples for each cluster.

\begin{figure}[htbp]
  \centering

  \begin{minipage}[c]{0.5\textwidth}
    \centering
    \includegraphics[width=\linewidth]{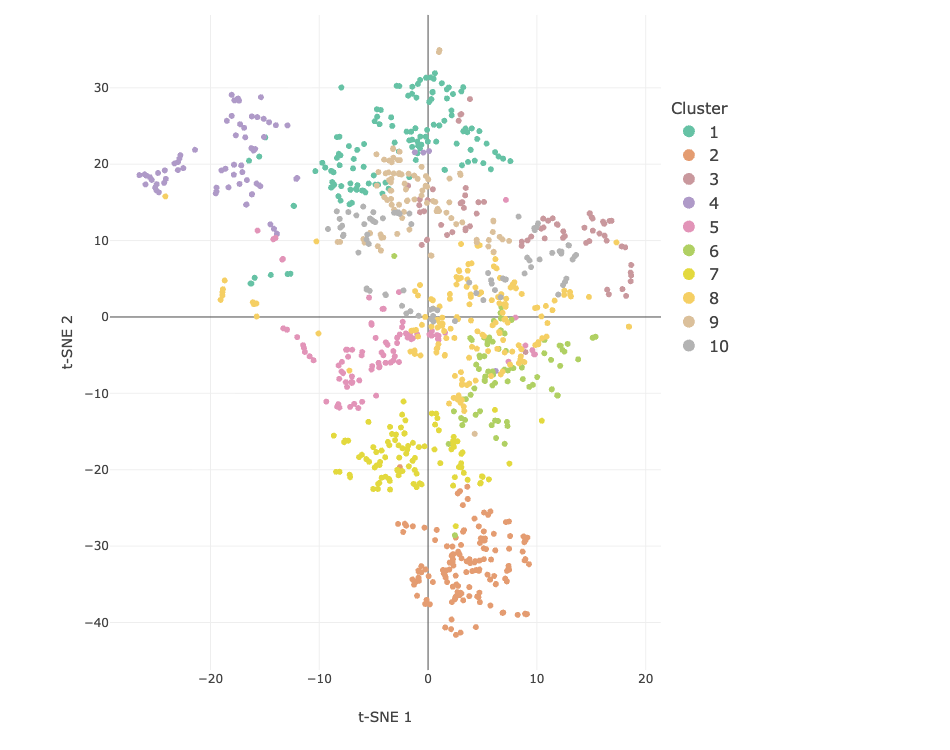}
  \end{minipage}%
  \begin{minipage}[c]{0.6\textwidth}
    \centering
    \includegraphics[width=\linewidth]{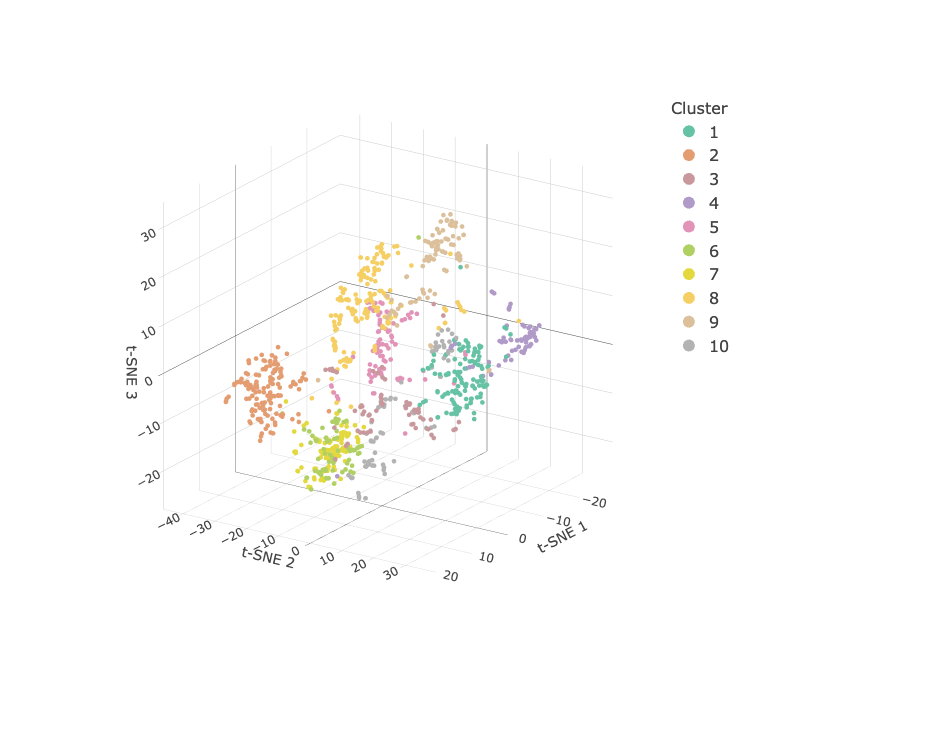}
  \end{minipage}

  \caption{Two- and three-dimensional t-SNE maps for Mandarin
  reduplicative constructions, with color highlighting by the clusters
  identified by the k-means algorithm. An interactive version is
  available in the anonymized supplementary materials.}
  \label{fig:tsne_cluster}
\end{figure}

The left panel of Figure~\ref{fig:tsne_cluster} presents the 10 groups in a two-dimensional plane obtained with t-distributed stochastic neighbor embedding (t-SNE; \citealt{Maatentsne}), a dimensionality-reduction technique for visualization. The t-SNE map shows that the 10 groups are remarkably well distinguishable even after reducing 200-dimensional embeddings onto a two-dimensional space.
The right panel of Figure~\ref{fig:tsne_cluster}
presents the results of a three-dimensional t-SNE clustering. Again, we find that the formations of the 10 clusters form good clusters.
These exploratory plots indicate  that Mandarin reduplicative constructions are not randomly distributed in the embedding space, but instead form remarkably differentiated clusters.  

\FloatBarrier

\begin{table}[htbp]
\centering
\caption{Misclassification table for LDA prediction of k-means clusters under leave-one-out cross-validation.}
\label{tab:LDA}
\small
\centering
\begin{tabular}[t]{rrrrrrrrrr}
\toprule
1 & 2 & 3 & 4 & 5 & 6 & 7 & 8 & 9 & 10\\
\midrule
122 & 0 & 2 & 3 & 0 & 0 & 0 & 2 & 0 & 3\\
0 & 134 & 0 & 0 & 0 & 1 & 2 & 0 & 0 & 0\\
3 & 0 & 64 & 0 & 0 & 2 & 0 & 2 & 1 & 1\\
3 & 0 & 0 & 63 & 1 & 1 & 0 & 3 & 0 & 0\\
3 & 0 & 0 & 1 & 80 & 0 & 0 & 5 & 1 & 1\\
\addlinespace
0 & 1 & 3 & 0 & 0 & 64 & 2 & 3 & 2 & 1\\
0 & 3 & 0 & 0 & 0 & 4 & 88 & 1 & 0 & 0\\
1 & 0 & 4 & 1 & 5 & 0 & 1 & 146 & 5 & 1\\
4 & 0 & 3 & 0 & 2 & 0 & 0 & 1 & 81 & 0\\
1 & 0 & 1 & 0 & 1 & 2 & 0 & 2 & 0 & 72\\
\bottomrule
\end{tabular}
\end{table}

\begin{table}[htbp]
\centering
\caption{Examples for each k-means cluster.}
\label{tab:cluster_examples}
\footnotesize
\setlength{\tabcolsep}{4pt}
\renewcommand{\arraystretch}{1.1}
\begin{tabularx}{\textwidth}{>{\centering\arraybackslash}p{1.1cm} >{\raggedright\arraybackslash}X >{\raggedright\arraybackslash}X}
\toprule
Cluster & Example 1 & Example 2 \\
\midrule
1 &
层层叠叠 \textit{ceng2-ceng2-die2-die2}, `layer upon layer' &
星星点点 \textit{xing1-xing1-dian3-dian3}, `in large amount but a scattered way' \\

2 &
讨论讨论 \textit{tao3-lun4-tao3-lun4}, `discuss a bit' &
放松放松 \textit{fang4-song1-fang4-song1}, `relax a bit' \\

3 &
进进出出 \textit{jin4-jin4-chu1-chu1}, `pass in and out' &
是是非非 \textit{shi4-shi4-fei1-fei1}, `rights and wrongs' \\

4 &
血红血红 \textit{xue4-hong2-xue4-hong2}, `intensely blood-red' &
黏黏糊糊 \textit{nian2-nian2-hu2-hu2}, `sticky and gooey' \\

5 &
开开心心 \textit{kai1-kai1-xin1-xin1}, `in a happy mood' &
顺顺当当 \textit{shun4-shun4-dang1-dang1}, `smooth and without trouble' \\

6 &
许许多多 \textit{xu3-xu3-duo1-duo1}, `a great many' &
很好很好 \textit{hen3-hao3-hen3-hao3}, `very good' \\

7 &
恭喜恭喜 \textit{gong1-xi3-gong1-xi3}, `many congratulations' &
客气客气 \textit{ke4-qi4-ke4-qi4}, `you are too polite' \\

8 &
哭哭啼啼 \textit{ku1-ku1-ti2-ti2}, `weep and wail, pitiful' &
温温柔柔 \textit{wen1-wen1-rou2-rou2}, `gentle and tender' \\

9 &
叮叮当当 \textit{ding1-ding1-dang1-dang1}, `clanging and jingling' &
摇摇晃晃 \textit{yao2-yao2-huang4-huang4}, `swaying and wobbling' \\

10 &
朝朝暮暮 \textit{zhao1-zhao1-mu4-mu4}, `from dawn to dusk, day after day' &
一步一步 \textit{yi1-bu4-yi1-bu4}, `step by step' \\
\bottomrule
\end{tabularx}
\end{table}

Table~\ref{tab:cluster_examples} shows two examples for each cluster. Some clusters appear to be relatively specialized. Cluster 2, for instance, is dominated by ABAB forms. The words in this cluster carry attenuative or delimitative verbal meanings, as illustrated by 讨论讨论 \textit{tao3-lun4-tao3-lun4}, `discuss a bit', derived from 讨论 \textit{tao3-lun4}, `discuss'.  Cluster 3 comprises mostly AABB formations, including both verbal and nominal formations. These formations bring opposites together, as in 进进出出 \textit{jin4-jin4-chu1-chu1}, `pass in and out', with 进出 \textit{jin4-chu1}, `pass in and out', as the base verb with approximately the same meaning.

Other clusters are more mixed. Cluster 6, for example, includes both AABB and ABAB forms. These reduplications are associated with quantitative increase, as in 许许多多 \textit{xu3-xu3-duo1-duo1}, `a great many', from 许多 \textit{xu3-duo1}, `many', and intensified degree, as in 很好很好 \textit{hen3-hao3-hen3-hao3}, `very good', from 很好 \textit{hen3-hao3}, `good'. 

Clusters 4 and 6 mainly comprise adjectives with intensified meanings, and cluster 5 contains many adverbial expressions.  Both verbs and adjectives are found in cluster 8. Cluster 9 brings together various onomatopoeic formations, but also comprises verbs such as 摇摇晃晃 \textit{yao2-yao2-huang4-huang4} `to be swaying and wobbling'.   Clusters 1 and 10 comprise formations expressing incremental and completive semantics. Cluster 7 consists entirely of ABAB forms and brings together expressions used in interpersonal exchange such as politeness routines, congratulations, and reassurance. Examples include 客气客气 \textit{ke4-qi4-ke4-qi4}, `you are too polite', and 恭喜恭喜 \textit{gong1-xi3-gong1-xi3}, `many congratulations'. These expressions typically are independent utterances by themselves. 

The cluster analysis shows that there is considerable structure in the semantic space of reduplicated formations. However, the cluster analysis remains  silent as to which properties underlie the observed clusters. In what follows, we therefore complement our informal characterization of the 10 clusters with a series of analyses that probe the factors that co-determine the meanings of the different reduplicative formations, and that give rise to the clusters detected by the k-means algorithm. We first examine the role of construction type (AABB/ABAB) and lexical category. We then turn to a series of non-categorical factors relating to valence, ego-relatedness, and sound-symbolism. 

\section{Semantic profiling}
\label{sec:semantic-profiling}
The previous section showed that reduplicative constructions are not randomly scattered in semantic space, and instead show considerable structure. This leads to the question of which grammatical and semantic properties underlie this structure. In this section, we profile the clusters along a set of linguistic properties in order to determine which properties characterize individual clusters and to what extent they differentiate the two reduplicative constructions. The linguistic properties that we consider range from relatively straightforward ones such as lexical category, to more gradient abstract  cognitive dimensions, such as valence, self-relatedness, and sound-relatedness. 

\subsection{Methods}
\subsubsection{Profiling method}
Our profiling procedure adapts the category-defining vector (CDV) approach introduced by \citet{westbury2019wordcategory, westbury2015emotion, westburywurm2022self}. The basic idea is that a semantic or grammatical property can be represented in semantic space by the averaged embeddings of a set of prototypical anchor words, i.e., the centroid of these embeddings. The  degree to which a given word instantiates that property  can be estimated from its position relative to the centroid. For a dimension $k$ with anchor set $A_k$, the CDV is defined as
\[
\mathbf{c}_k = \frac{1}{|A_k|} \sum_{a \in A_k} \mathbf{v}_a
\]
where $A_k$ is the anchor set for property $k$, and $\mathbf{v}_a$ is the embedding vector of anchor word $a$.

For a unipolar property such as self-relatedness, the profile score of a reduplicative construction is defined as the cosine similarity between its embedding vector and the corresponding centroid (CDV):
\[
s(r,k) = \cos(\mathbf{v}_r, \mathbf{c}_k)
\]
where $\mathbf{v}_r$ is the embedding vector of reduplicative construction $r$. Higher values indicate a stronger association with the relevant property.

For bipolar properties, such as pairs of opposite emotions, we first constructed a semantic axis from the difference between two opposing centroids and then projected each reduplicative construction onto that axis:
\[
\mathbf{a}_{i,j} = \mathbf{c}_i - \mathbf{c}_j,
\qquad
p(r;i,j) = \mathbf{v}_r \cdot \frac{\mathbf{a}_{i,j}}{\|\mathbf{a}_{i,j}\|}
\]
Positive and negative values indicate closer alignment with one or the other pole of the corresponding property.

\subsubsection{Anchor-word selection}

For each profiling property, we selected 100 high-frequency anchor words that served as prototypical representatives of the semantic pole or category of interest. Emotion-related anchor words were selected with reference to the Chinese Emotional Lexicon Ontology \citep{xu2008affective}. Lexical category information was determined with reference to \textit{A Frequency Dictionary of Mandarin Chinese} \citep{XiaoRaysonMcEnery2009}, which provides frequency-based lexical entries together with part-of-speech indexing.  For selecting anchors, we ordered candidate words by decreasing frequency, and selected the most frequent candidates that most closely and most clearly represented a given prototypical pole. Furthermore, only those anchor words were included for which an embedding was available in the Tencent resource. 

\subsection{Construction: AABB vs. ABAB}

The two construction types show markedly different distributions across the clusters, as shown in Table~\ref{tab:pattern_distribution}. While some clusters are strongly associated with AABB, others are dominated by ABAB. For example, Cluster 5 contains only AABB items, whereas Cluster 7 contains only ABAB items. Clusters 2, 6, and 8 also show a strong bias toward one construction type. These patterns suggest that construction type is not randomly distributed across the semantic space.

\begin{table}[htbp]
\centering
\caption{Distribution of AABB and ABAB patterns across the 10 clusters.}
\label{tab:pattern_distribution}
\small
\begin{tabular}{rrrrrr}
\toprule
Cluster & AABB & \% & ABAB & \% & Total \\
\midrule
1  & 88 & 66.67  & 44 & 33.33  & 132 \\
2  & 1  & 0.73    & 136 & 99.27 & 137 \\
3  & 61 & 83.56  & 12 & 16.44  & 73 \\
4  & 47 & 66.20  & 24 & 33.80  & 71 \\
5  & 91 & 100  & 0 & 0     & 91 \\
6  & 5  & 6.58    & 71 & 93.42  & 76 \\
7  & 0  & 0     & 96 & 100  & 96 \\
8  & 150 & 91.46 & 14 & 8.54   & 164 \\
9  & 35 & 38.46  & 56  & 61.54  & 91 \\
10 & 42 & 53.16  & 37 & 46.84  & 79 \\
\bottomrule
\end{tabular}
\end{table}

If the two patterns encode partly distinct constructional meanings, construction types should be predictable from the embeddings. To address this issue, we trained a linear discriminant analysis (LDA) classifier using the embedding vectors as predictors and construction type (AABB vs.\ ABAB) as the response variable. Model performance was evaluated using leave-one-out cross-validation. The classifier achieved an accuracy of 96.44\%, substantially above the majority-class baseline of 51.49\%, with 506 of 520 AABB items and 468 of 490 ABAB items correctly identified.

The LDA analysis shows that the two reduplication patterns are much more separable in the semantic space than expected on the basis of the counts of construction types within clusters presented in Table~\ref{tab:pattern_distribution}. One cluster comprises only AABB constructions (cluster 5) and one cluster (cluster 7) has only ABAB constructions.  The other clusters mix the two constructions to different degrees.  This suggests that the clusters are not only determined by construction pattern, but by many other linguistic predictors as well.  One obvious candidate is lexical category.

\subsection{Lexical category}
Table~\ref{tab:lexical_distribution} presents the counts of verbs, nouns, adjectives, and adverbs, broken down by cluster. Nouns don't show any clear preference for clusters, verbs favor clusters 2 and 7, adjectives favor clusters 4, 5 and 8, and adverbs favor clusters 9 and 10. 
This distributional differentiation is supported by a linear discriminant analysis (LDA) using embeddings to predict lexical category. Under leave-one-out cross-validation, the model achieved an accuracy of 84.06\%, well above the majority-class baseline of 45.05\%, correctly classifying 405 of 455 adjectives, 154 of 199 adverbs, 22 of 33 nouns, and 268 of 323 verbs.

However, this discrete way of assigning lexical category is rather coarse, as in Mandarin Chinese, a word can be used across many different lexical categories. For example, `明明白白' \textit{ming2-ming2-bai2-bai2} `clear; clearly' can be used in a more adjectival way, as in `大家都明明白白' \textit{da4-jia1 dou1 ming2-ming2-bai2-bai2} `everyone understands it clearly', or in a more adverbial way, as in `他曾明明白白地吩咐过' \textit{ta1 ceng2 ming2-ming2-bai2-bai2 de fen1-fu4 guo4} `he had once instructed [someone] very clearly', depending on context.\footnote{Both examples are drawn from the CCL corpus used in this study.} 

Therefore, we constructed a Noun-Verb axis and an Adjective-Adverb axis. We then projected all reduplicative constructions onto these two continuous lexical axes, to address whether particular regions of the reduplicative semantic space show stronger noun-like, verb-like, adjective-like, or adverb-like tendencies. 

\begin{table}[htbp]
\centering
\caption{Distribution of lexical categories across the 10 clusters.}
\label{tab:lexical_distribution}
\small
\begin{tabular}{rrrrrrrrrr}
\toprule
Cluster & Noun & \% & Verb & \% & Adjective & \% & Adverb & \% & Total \\
\midrule
1  & 12 & 9.09  & 4   & 3.03  & 71  & 53.79 & 45 & 34.09 & 132 \\
2  & 0  & 0   & 134 & 97.81 & 3   & 2.19  & 0  & 0   & 137 \\
3  & 3  & 4.11  & 28  & 38.36 & 33  & 45.21 & 9  & 12.33 & 73 \\
4  & 2  & 2.82  & 4   & 5.63  & 62  & 87.32 & 3  & 4.23  & 71 \\
5  & 2  & 2.20  & 0   & 0   & 88  & 96.70 & 1  & 1.10  & 91 \\
6  & 2  & 2.63  & 19  & 25  & 42  & 55.26 & 13 & 17.11 & 76 \\
7  & 2  & 2.08  & 73  & 76.04 & 13  & 13.54 & 8  & 8.33  & 96 \\
8  & 1  & 0.61  & 50  & 30.49 & 105 & 64.02 & 8  & 4.88  & 164 \\
9  & 0  & 0   & 10  & 10.99 & 16  & 17.58 & 65 & 71.43 & 91 \\
10 & 9  & 11.39 & 1   & 1.27  & 22  & 27.85 & 47 & 59.49 & 79 \\
\bottomrule
\end{tabular}
\end{table}

\begin{figure}[htbp]
\centering
\begin{minipage}[t]{0.47\linewidth}
    \centering
    \includegraphics[width=\linewidth, trim=100 100 100 100, clip]
    {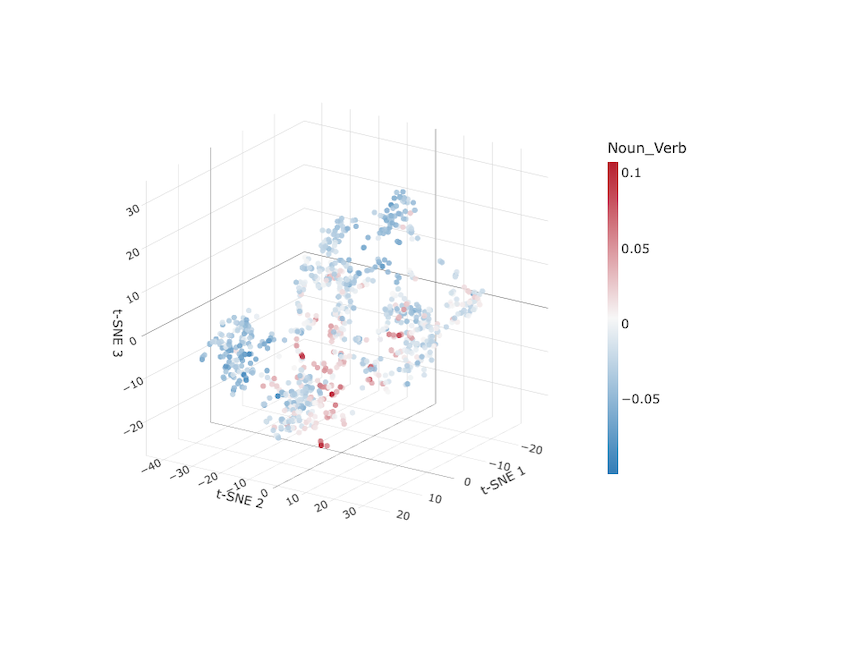}
\end{minipage}
\hspace{0.02\linewidth}
\begin{minipage}[t]{0.47\linewidth}
    \centering
    \includegraphics[width=\linewidth, trim=100 100 100 100, clip]
    {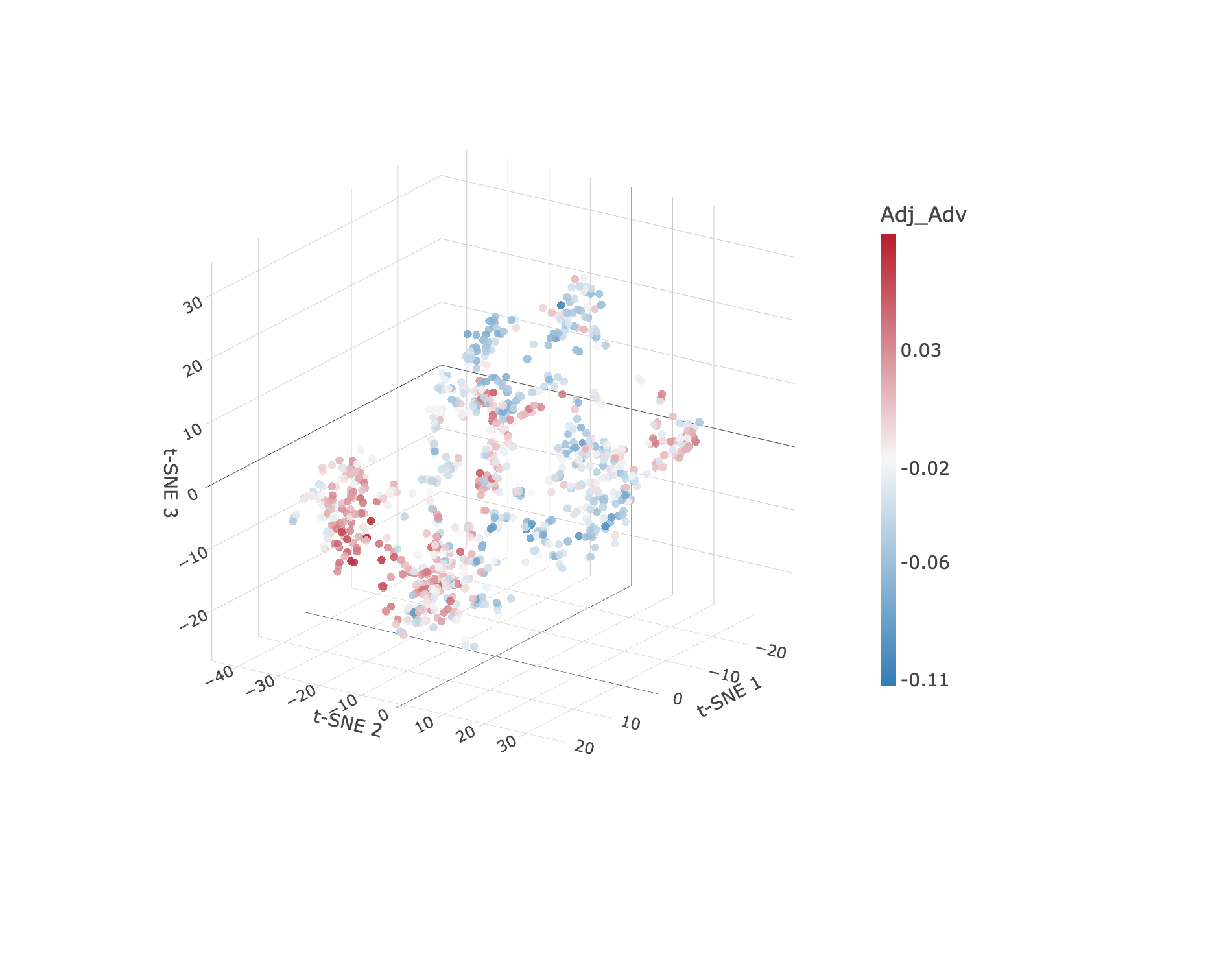}
\end{minipage}
  \caption{Three-dimensional t-SNE maps of Mandarin reduplicative constructions, with color highlighting for lexical category along the Noun-Verb axis (left panel) and the Adjective-Adverb axis (right panel). Darker red dots indicate more noun-like and adjective-like items, whereas deeper blue dots indicate more verb-like and more adverb-like items. An interactive version is available in the anonymized supplementary materials.}
  \label{fig:lexical_category}
\end{figure}

Figure~\ref{fig:lexical_category} shows that, as expected, lexical categories are not randomly distributed across the semantic space. In the left panel, Clusters 10, 5, and 1 show relatively stronger noun-like tendencies. Illustrative examples from these clusters are 方方面面 \textit{fang1-fang1-mian4-mian4} `all aspects' in Cluster 10, 山山水水 \textit{shan1-shan1-shui3-shui3} `hills and rivers' in Cluster 1, and 条条框框 \textit{tiao2-tiao2-kuang4-kuang4} `rules and regulations' in Cluster 5. 
Clusters 6 and 2 are relatively more verb-like, with some of the darkest blue datapoints corresponding to items such as 选择选择 \textit{xuan3-ze2-xuan3-ze2} `choose a bit' in Cluster 6 and 请教请教 \textit{qing3-jiao4-qing3-jiao4} `consult briefly' in Cluster 2.

A similar distribution can be observed on the Adjective-Adverb axis, shown in the right panel of Figure~\ref{fig:lexical_category}. The left part of the semantic space, especially Clusters 5 and 6, is more adjective-like, with more red  dots such as 轰轰烈烈 \textit{hong1-hong1-lie4-lie4} `vigorous and dynamic' in Cluster 5 and 很贵很贵 \textit{hen3-gui4-hen3-gui4} `very expensive indeed' in Cluster 6. By contrast, adverb-like tendencies are found mainly in Cluster 3 and, to a lesser extent, in Cluster 6, represented by deeper blue datapoints such as 前前后后 \textit{qian2-qian2-hou4-hou4} `back and forth' in Cluster 3 and 许久许久 \textit{xu3-jiu3-xu3-jiu3} `for a very long time' in Cluster 6.

We also found that lexical categories are related to reduplicative pattern. More verb-like tendencies are mainly associated with ABAB constructions, whereas more noun-like tendencies are mainly associated with AABB constructions. Adjectival and adverbial tendencies are attested in both patterns, but both are more common in AABB constructions and less frequent in ABAB constructions.

To assess the extent to which our gradient measures for lexical categories contribute to cluster structure, we used the Noun-Verb and Adjective-Adverb axis scores to predict cluster, using linear discriminant analysis (LDA). Under leave-one-out cross-validation, the model achieved an accuracy of 32.97\% (majority-class baseline: 16.24\%). These results suggest that lexical category contributes to the organization of the cluster structure, but does not fully account for it. Some clusters, especially Clusters 2 and 8, were identified with moderate success, whereas others, such as Clusters 3, 4, and 6, were not recoverable from the lexical-category scores alone (see Table~\ref{tab:confusionClusterAxes}). This in turn raises the possibility that part of the remaining structure is related to more abstract semantic properties. In what follows, we therefore extend the profiling analysis to dimensions such as valence, self-relatedness, and sound-related meaning.

\begin{table}[H]
\centering
\caption{Accuracies by cluster for semantic-space profiling across four categories.}
\label{tab:confusionClusterAxes}
\small
\begin{tabular}{rrrrr}
\toprule
Cluster & Lexical category & Valence & Ego & Sound-related meaning \\
\midrule
1  & 0.136 & 0.500 & 0.803 & 0.765 \\
2  & 0.715 & 0.606 & 0.212 & 0.672 \\
3  & 0.000 & 0.000 & 0.000 & 0.356 \\
4  & 0.000 & 0.000 & 0.000 & 0.085 \\
5  & 0.429 & 0.440 & 0.000 & 0.264 \\
6  & 0.000 & 0.092 & 0.000 & 0.013 \\
7  & 0.083 & 0.073 & 0.677 & 0.646 \\
8  & 0.665 & 0.274 & 0.512 & 0.543 \\
9  & 0.286 & 0.275 & 0.000 & 0.637 \\
10 & 0.443 & 0.000 & 0.000 & 0.468 \\
\bottomrule
\end{tabular}
\end{table}

\subsection{Valence}
\begin{figure}[htbp]
  \centering
  \begin{minipage}{0.32\linewidth}
    \includegraphics[width=\linewidth, trim=240 280 280 280, clip]{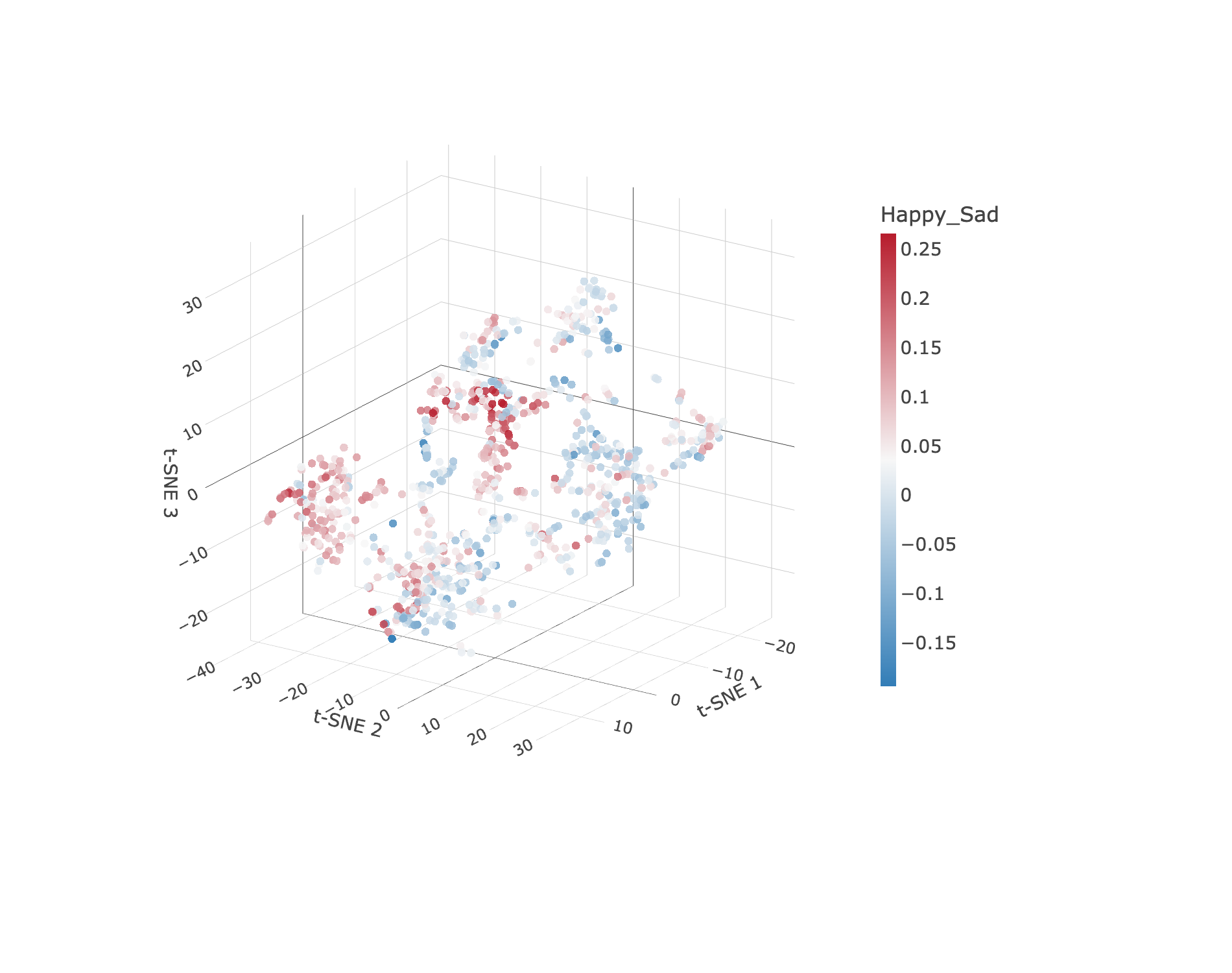}
  \end{minipage}
  \hfill
  \begin{minipage}{0.32\linewidth}
    \includegraphics[width=\linewidth, trim=240 280 280 280, clip]{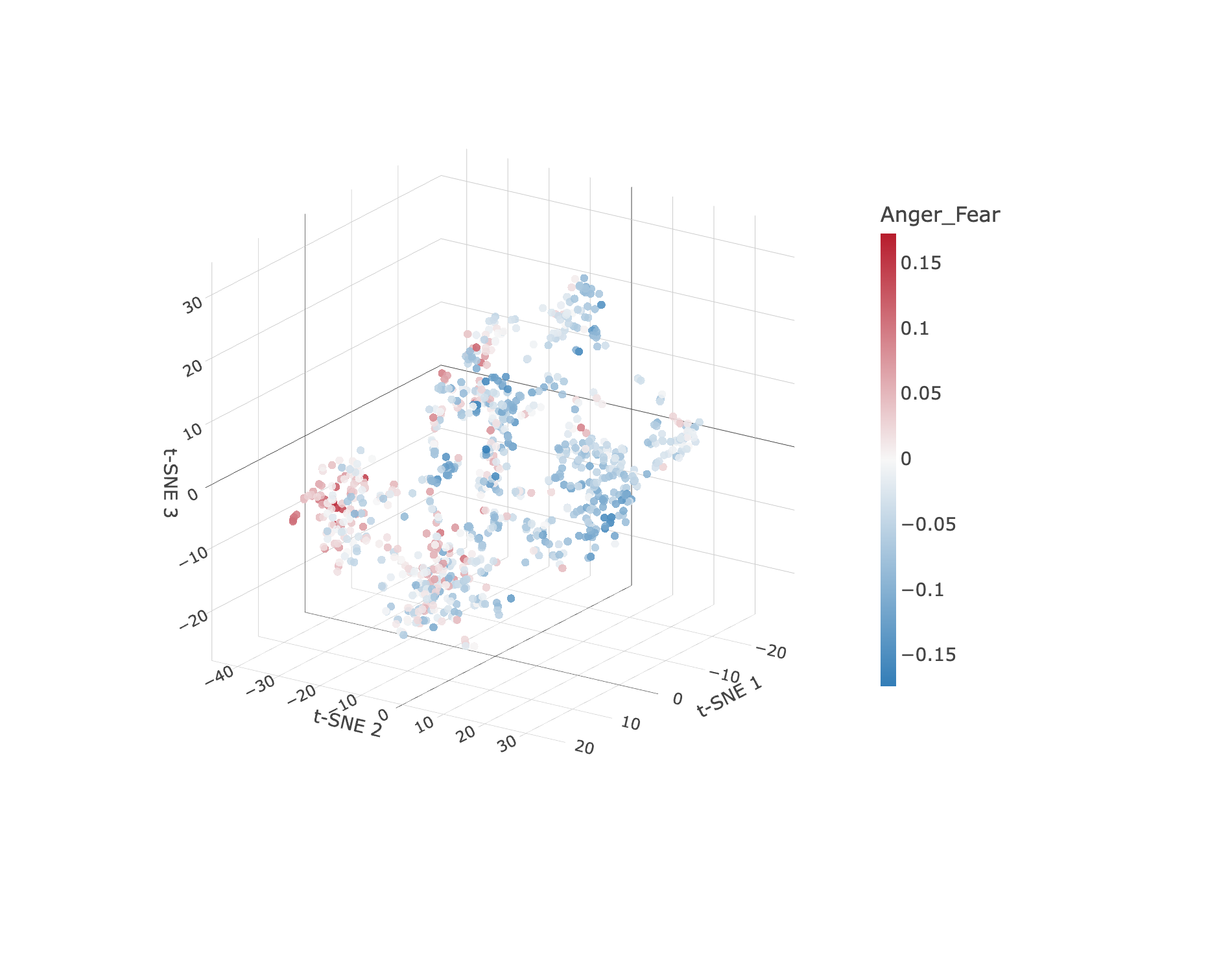}
  \end{minipage}
  \hfill
  \begin{minipage}{0.32\linewidth}
    \includegraphics[width=\linewidth, trim=240 280 260 280, clip]{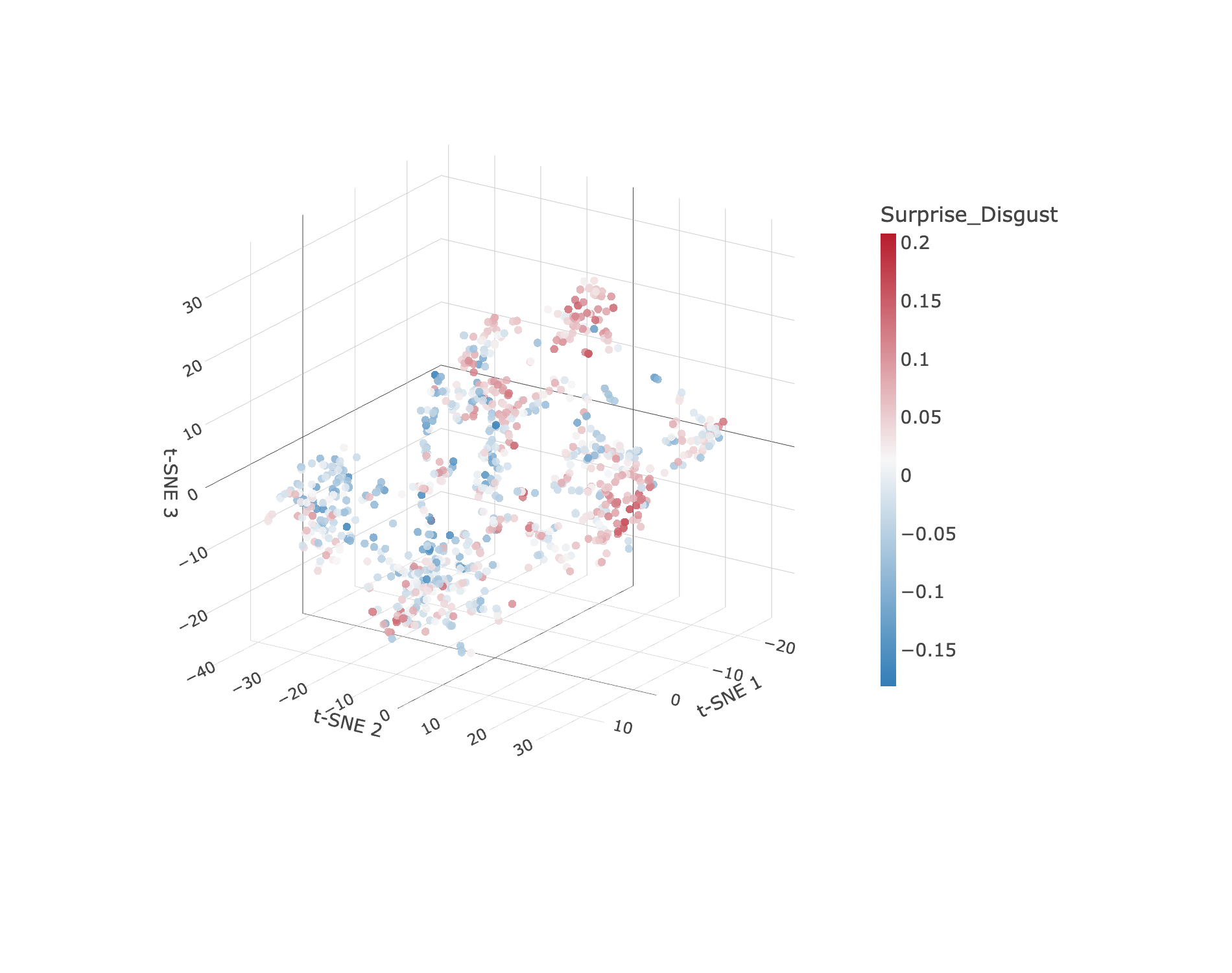}
  \end{minipage}
  \caption{Three-dimensional t-SNE maps of Mandarin reduplicative constructions, with color highlighting valence along the Happy-Sad axis (left panel), Anger-Fear axis (middle panel), and Surprise-Disgust axis (right panel). Darker red dots indicate higher degrees of happiness, anger and surprise, whereas darker blue dots indicate higher degrees of sadness, fear and disgust. An interactive version is available in the anonymized supplementary materials.}
  \label{fig:valence}
\end{figure}

For valence profiling, we selected six basic emotions \citep{Ekman1992} and organized them along three contrastive axes: Happy-Sad, Anger-Fear and Surprise-Disgust. We then projected all reduplicative constructions onto these three axes, resulting in three scores for each formation.  The three panels of Figure~\ref{fig:valence} present the 3-D t-SNE space, with color highlighting for the three valence contrasts. Higher scores for happiness, anger, and surprise are presented with darker red colors, whereas more negative scores, indicating closer proximity to sadness, fear, and disgust, are represented by deeper shades of blue.  Figure~\ref{fig:valence} shows that different regions of the semantic space are associated with different aspects of emotion.

In the left panel, reduplications that score high on the happy-sad axis tend to cluster in the left-central part of the semantic space, especially in Clusters 5, 2, and 7, with reduplications such as 开开心心 \textit{kai1-kai1-xin1-xin1} `very happy' in Cluster 5, 高兴高兴 \textit{gao1-xing4-gao1-xing4} `cheer up a bit' in Cluster 2, and 祝贺祝贺 \textit{zhu4-he4-zhu4-he4} `many congratulations' in Cluster 7 (for cluster identifiers, compare Figure~\ref{fig:tsne_cluster}). Reduplications that score low on the happy-sad axis are more common in the middle and right parts of the semantic space, especially in Clusters 8, 6, and 9. Among the more extreme formations we find 悲悲切切 \textit{bei1-bei1-qie4-qie4} `deeply sorrowful' in Cluster 8, 很痛很痛 \textit{hen3-tong4-hen3-tong4} `very painful' in Cluster 6, and 滴滴答答 \textit{di1-di1-da1-da1} `dripping continuously' in Cluster 9.

In the middle panel, the reduplications that score high on the anger-fear axis are more common on the left side of the semantic space, especially in Clusters 8 and 2, with items such as 教训教训 \textit{jiao4-xun4-jiao4-xun4} `teach someone a lesson' and 骂骂咧咧 \textit{ma4-ma4-lie1-lie1} `to swear while talking'. Reduplications that score low on the anger-fear axis, by contrast, are more common on the right side of the t-SNE space, including 紧紧张张 \textit{jin3-jin3-zhang1-zhang1} `in a highly strained state' in Cluster 8, 迷迷茫茫 \textit{mi2-mi2-mang2-mang2} `all lost and bewildered' in Cluster 1, and 颠颠簸簸 \textit{dian1-dian1-bo3-bo3} `bumping and jolting all the way' in Cluster 3.

The third panel shows the results for the scores on the surprise-disgust axis. Words with higher scores are more common in the right-hand part of the semantic space, and include formations such as 飘飘渺渺 \textit{piao1-piao1-miao3-miao3} `floating and indistinct' (Cluster 1), 扑通扑通 \textit{pu1-tong1-pu1-tong1} `pounding rapidly' (Cluster 9), and 恭喜恭喜 \textit{gong1-xi3-gong1-xi3} `many congratulations' (Cluster 7). Why 飘飘渺渺 scores so high on this axis is unclear to us. Reduplications with lower scores, indicating closer alignment with the disgust pole of the axis, are concentrated on the left side of the semantic space, especially in Cluster 8. Among them we have 讨厌讨厌 \textit{tao3-yan4-tao3-yan4} `very annoying' and 邋邋遢遢 \textit{la1-la1-ta1-ta1} `sloppy and untidy'.

To assess whether the scores on these three emotional axes predict cluster membership, we again used LDA. Under leave-one-out cross-validation, the model achieved an accuracy of 27.03\% (majority-class baseline: 16.24\%). This indicates that these emotional axes are somewhat related to cluster structure, but can only provide a partial explanation. Some clusters, especially Clusters 1, 2, and 5, were recovered with moderate success, whereas others, such as Clusters 3, 4, and 10, were not recoverable from the emotion scores alone (cf. Table~\ref{tab:confusionClusterAxes}).

\subsection{Self-relatedness}
\begin{figure}[H]
  \centering
  \includegraphics[width=0.5\linewidth, trim=160 160 160 160, clip]{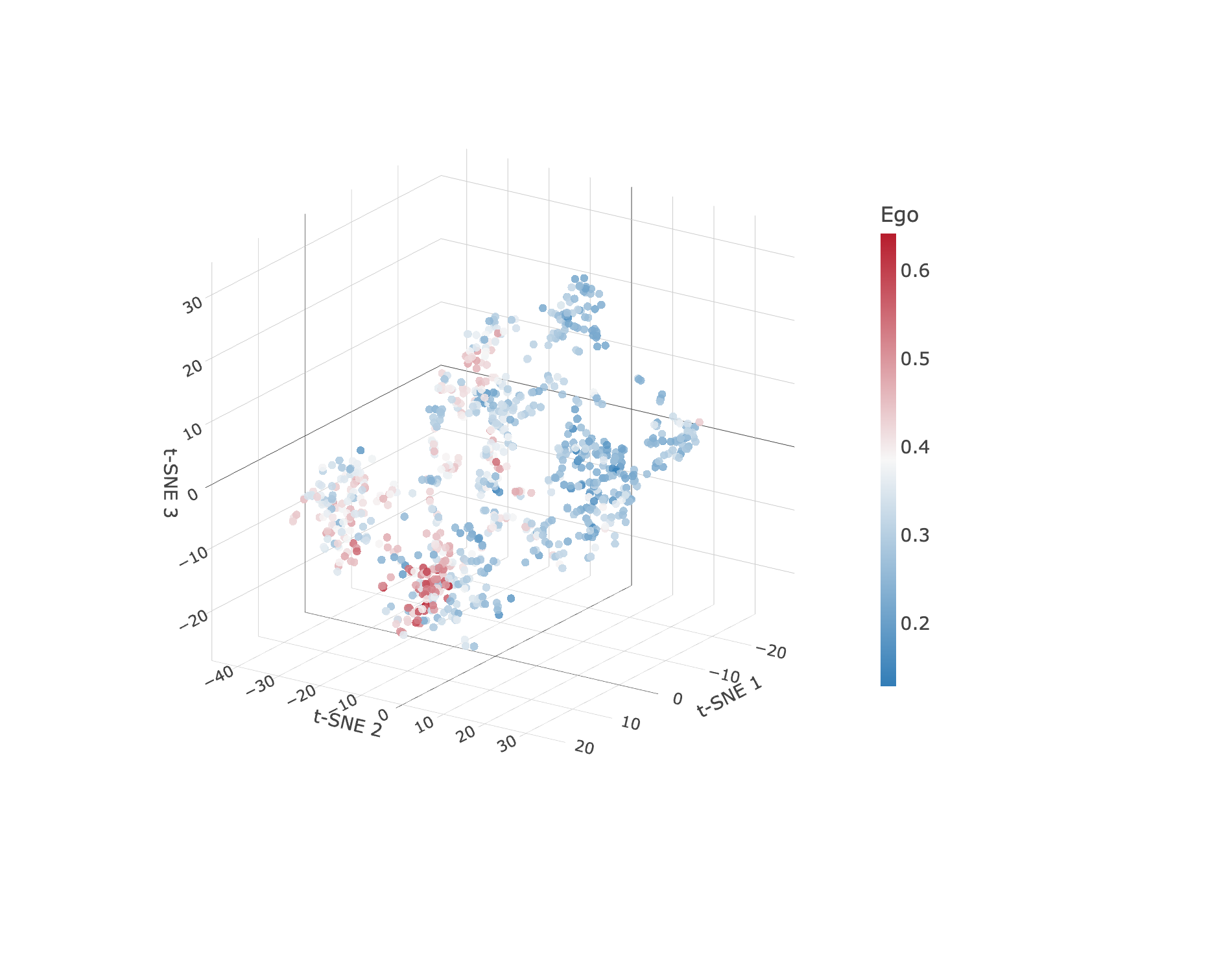}
  \caption{Three-dimensional t-SNE map of Mandarin reduplicative constructions, with color highlighting the cosine similarity with the centroid of self-related anchor words. Deeper shades of red represent higher cosine similarities, deeper shades of blue near orthogonality. An interactive version is available in the anonymized supplementary materials.}
  \label{fig:ego_tsne}
\end{figure}

To investigate self-relatedness, we calculated the cosine similarity of the reduplications with the centroid of the anchor words for the ``self''. Figure~\ref{fig:ego_tsne} highlights words with higher scores for the self with darker shades of red.  Cluster 7 shows particularly strong scores for the `self', such as 好说好说 \textit{hao3-shuo1-hao3-shuo1} `no problem', 罢了罢了 \textit{ba4-le5-ba4-le5} `let it go', and 多谢多谢 \textit{duo1-xie4-duo1-xie4} `many thanks'. These constructions are often used independently in interaction to express the speaker's own stance or attitude toward a situation. Examples of formations that are orthogonal to the `self' are 开开停停	(Cluster 3) \textit{kai1-kai1-ting2-ting2} `stop and go' and 角角落落 (Cluster 1) \textit{jiao3-jiao3-luo4-luo4} `all the corners' and 甜甜美美 (Cluster 5) \textit{tian2-tian2-mei3-mei3} `sweet and pleasant'. 

We also used ego scores to predict cluster membership with a linear discriminant analysis (LDA). Under leave-one-out cross-validation, the model achieved an accuracy of 28.12\% (majority-class baseline: 16.24\%). The effect is especially visible for Cluster 7, which also contains 19 of the 20 reduplications with the highest ego scores, but 6 out of 10 clusters are  not recoverable at all from ego scores alone (cf. Table~\ref{tab:confusionClusterAxes}).

\subsection{Sound-related meaning}
Manual inspection suggests that Cluster 9 is the cluster most strongly associated with onomatopoeic meaning in our data. To examine more systematically what kinds of sounds are encoded by reduplications, we drew on a classification of everyday sound sources and distinguished three broad categories:  vibrating solids, gases, and liquids \citep{gaver1993sound}. Since many reduplications in our data also reflect sounds  made by humans, we added a fourth category, human sounds.

\begin{figure}[H]
  \centering
  \begin{minipage}{0.48\linewidth}
    \centering
    \includegraphics[width=\linewidth, trim=160 160 160 160, clip]{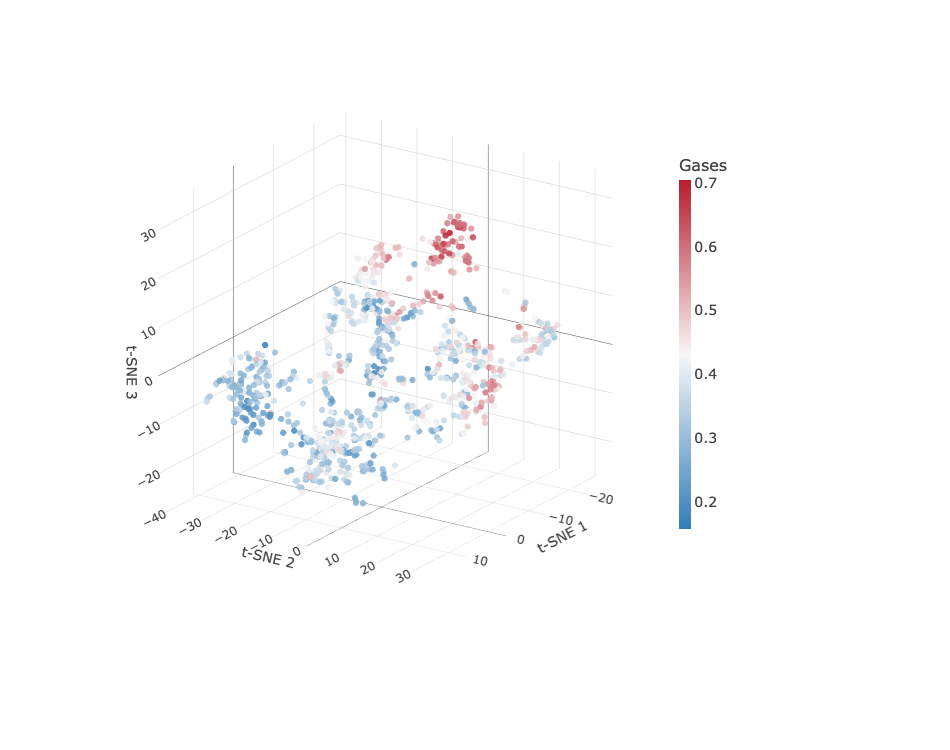}
  \end{minipage}
  \hfill
  \begin{minipage}{0.48\linewidth}
    \centering
    \includegraphics[width=\linewidth, trim=160 160 160 160, clip]{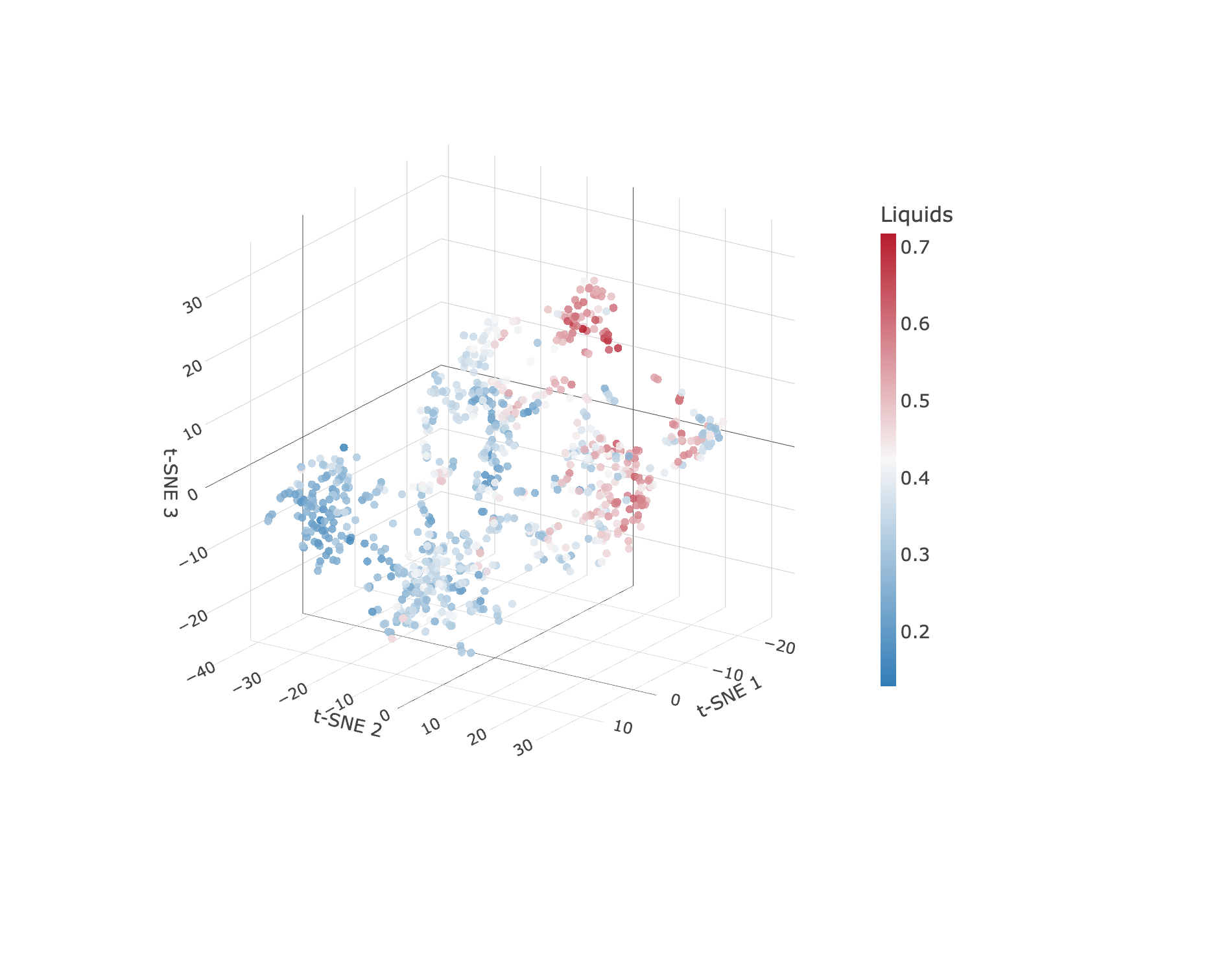}
  \end{minipage}

  \vspace{0.5em}

  \begin{minipage}{0.48\linewidth}
    \centering
    \includegraphics[width=\linewidth, trim=160 160 160 160, clip]{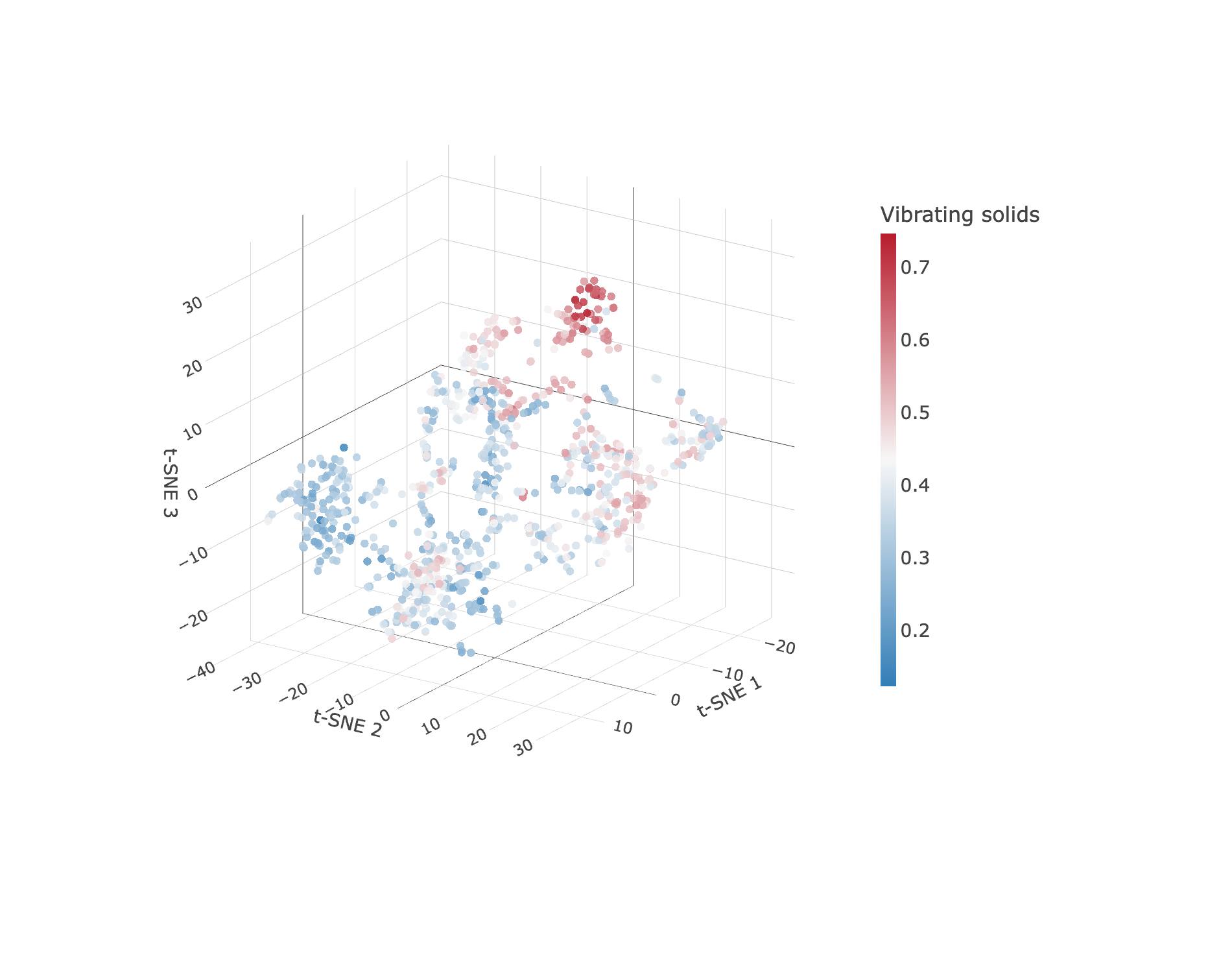}
  \end{minipage}
  \hfill
  \begin{minipage}{0.48\linewidth}
    \centering
    \includegraphics[width=\linewidth, trim=160 160 160 160, clip]{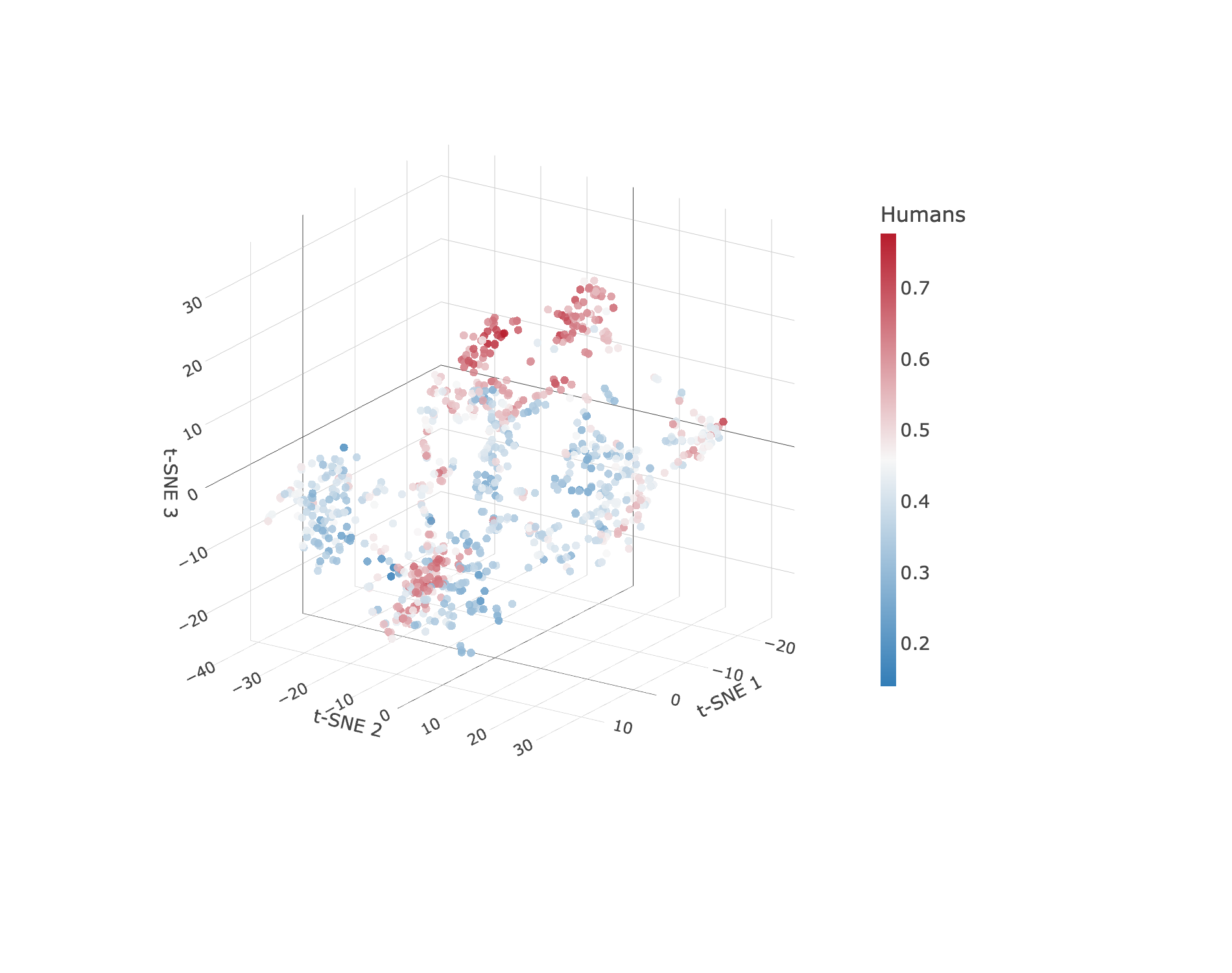}
  \end{minipage}
  \caption{Three-dimensional t-SNE maps of Mandarin reduplicative constructions, with color highlighting the cosine similarity with the centroid of anchor words of Gases (top left panel), Liquids (top right panel), Vibrating Solids (bottom left panel), and Humans (bottom right panel) categories. Deeper shades of red represent higher cosine similarities, deeper shades of blue near orthogonality. An interactive version is available in the anonymized supplementary materials.}
  \label{fig:sound}
\end{figure}

Figure~\ref{fig:sound} shows that three sound-source categories, gases, liquids, and vibrating solids, display broadly similar spatial tendencies, with darker shades of red concentrated especially in the upper-right part of the semantic space, where Cluster 9, Cluster 1, and Cluster 8 are located. 
Among them, Cluster 9 is especially associated with the gases category, with reduplications such as 轰隆轰隆 \textit{hong1-long1-hong1-long1} `continuous rumble' and 一股一股 \textit{yi4-gu3-yi4-gu3} `in waves', typically said of hot air. Examples of formations that are orthogonal to the gases are 圆圆满满 (Cluster 5) \textit{yuan2-yuan2-man3-man3} `fully satisfactory' and 活络活络 (Cluster 2) \textit{huo2-luo4-huo2-luo4} `loosen up a bit'.

Liquid-related formations are predominant in Clusters 9 and 1, and include formations such 哗啦哗啦 \textit{hua1-la1-hua1-la1} `a splashing sound' in Cluster 9 and 淅淅沥沥 \textit{xi1-xi1-li4-li4} `dripping lightly' in Cluster 1. 
Examples of reduplications that are orthogonal to the category of liquids are found in Cluster 2 and Cluster 5, such as 宣传宣传 \textit{xuan1-chuan2-xuan1-chuan2} `give wide publicity' and 体体面面 \textit{ti3-ti3-mian4-mian4} `very decent'.

Cluster 9 also contains formations relating to vibrating solids, with items such as 叮叮当当 \textit{ding1-ding1-dang1-dang1} `jingling of bells' and 咔嚓咔嚓 \textit{ka1-cha1-ka1-cha1} `cracking or snapping sounds'. Examples of reduplications that are orthogonal to the vibrating solids category are found in Cluster 10 and Cluster 2, including examples such as 方方面面 \textit{fang1-fang1-mian4-mian4} `all aspects' and 帮助帮助 \textit{bang1-zhu4-bang1-zhu4} `give some help'.

The human sound category shows a more extensive distribution than the three physical sound-source categories. In addition to the upper-right region, the middle and left parts of the semantic space also show shades of dark red, especially in Cluster 7 and Cluster 8. Cluster 7 is mainly associated with vocal expressions, with items such as 哎呀哎呀 \textit{ai1-ya1-ai1-ya1} `repeated exclamations of surprise or concern', compare to ``oh dear, oh dear'' in English. By contrast, Cluster 8 is more closely associated with sounds or sound-related effects of human action, as in 哆哆嗦嗦 \textit{duo1-duo1-suo1-suo1} `trembling and shivering'.
Examples of reduplications that are orthogonal to the human-sound category are found in Cluster 10 and Cluster 6, such as 事事物物 \textit{shi4-shi4-wu4-wu4} `all things' and 选择选择 \textit{xuan3-ze2-xuan3-ze2} `do some choosing'.

We also used the Gases, Liquids, Vibrating Solids, and Humans scores to predict clusters using LDA. Under leave-one-out cross-validation, the model achieved an accuracy of 49.11\% (majority-class baseline: 16.24\%). In particular, Cluster 7, Cluster 8, and Cluster 9 were recovered with moderate to high accuracy, as shown in Table~\ref{tab:confusionClusterAxes}.

\subsection{Predictability of clusters from semantic profiling}
The preceding sections profiled the semantic space starting from four categories: lexical category, valence, self-relatedness, and sound-related meaning. We next examine how well the semantic space can be explained when all semantic profiling categories are considered together. Therefore, we used all semantic categories, combined with construction type (AABB/ABAB) in an LDA analysis, asking the LDA to predict cluster identity. Under leave-one-out cross-validation, the model achieved an accuracy of 71.78\%, which is substantially above the majority-class baseline of 16.24\% (see Table~\ref{tab:LDA4Semantic}).

\begin{table}[htbp]
\centering
\caption{Misclassification table for LDA of k-means clusters from four semantic categories, using leave-one-out cross-validation. }
\label{tab:LDA4Semantic}
\small
\centering
\begin{tabular}{rrrrrrrrrrrr}
  \hline
 & 1 & 2 & 3 & 4 & 5 & 6 & 7 & 8 & 9 & 10 & by-row accuracy \\ 
  \hline
1 & 98  & 0  & 7  & 14  & 1  & 1  & 0  & 1  & 13  & 3  & 0.71 \\ 
  2 & 0  & 130  & 0  & 0  & 1  & 10  & 5  & 0  & 0  & 1  & 0.88 \\ 
  3 & 12  & 0  & 31  & 3  & 0  & 1  & 0  & 6  & 1  & 10  & 0.48 \\ 
  4 & 7  & 0  & 3  & 21  & 2  & 1  & 1  & 4  & 3  & 0  & 0.50 \\ 
  5 & 1  & 1  & 2  & 6  & 80  & 1  & 0  & 16  & 0  & 6  & 0.71 \\ 
  6 & 2  & 2  & 5  & 4  & 0  & 46  & 8  & 6  & 2  & 4  & 0.58 \\ 
  7 & 0  & 4  & 1  & 1  & 0  & 5  & 81  & 1  & 0  & 0  & 0.87 \\ 
  8 & 3  & 0  & 9  & 12  & 6  & 2  & 1  & 122  & 7  & 1  & 0.75 \\ 
  9 & 1  & 0  & 4  & 10  & 0  & 1  & 0  & 7  & 65  & 3  & 0.71 \\ 
  10 & 8  & 0  & 11  & 0  & 1  & 8  & 0  & 1  & 0  & 51  & 0.64 \\ 
   \hline
\end{tabular}
\end{table}

Several clusters were predicted with high accuracy, including Cluster 2 (0.88), Cluster 7 (0.87), and Cluster 1 (0.71), although others, especially Cluster 3 and Cluster 4, are not well recoverable. 

In summary, reduplications show considerable clustering in semantic space. A k-means clustering suggested 10 clusters that are well supported by an LDA analysis predicting cluster from embeddings.  We have shown that cluster identity can be predicted to a considerable extent (and far above majority baselines) from a range of linguistic features: reduplication construction type, word category, valency, self-relevance, and type of sound.  Nevertheless, the individual clusters are not fully defined by these linguistic features, from which we conclude that there is further systematic variation in the semantic space that resists reduction to the linguistic variables we investigated.

In the next section, we turn to the question of to what extent reduplicative meanings are related to the meanings of their base words and, if so, how this relation can be characterized.

\section{Shift vectors}
\label{sec:shift-vectors}

In distributional-semantic research on morphology, shift vectors have been used to represent the change in meaning from a base form to a morphologically more complex form \citep[see, e.g.,][]{drozd2016word,boleda2020distributional,mikolov2013linguistic}. This approach has proved useful for a range of phenomena, including English noun plurals \citep{ShafaeiBajestanUhrigBaayen2022}, Finnish noun inflection \citep{NikolaevEtAl2022}, Russian nominal paradigms \citep{ChuangEtAl2022}, German particle verbs and derivation \citep{StupakBaayen2022}, and Mandarin suffixation \citep{ShenBaayen2022b}. Across these studies, shift vectors have been used to ask whether a change in morphological form (e.g., realized with suffixation) goes hand in hand with a change in embedding space. If a morphological meaning change is systematic, words sharing the same form structure are expected to show similar displacement in semantic space. In the present study, we apply this approach to Mandarin reduplication, analyzing how reduplicative meanings relate to the meanings of their base words. Of particular interest is whether the semantic change from base words to reduplicative constructions is systematically different across the different kinds of reduplicative constructions.

Let $\bm{v}_r$ denote the vector of a reduplication and $\bm{v}_b$ the vector of its base word. The semantic shift $\bm{s}_r$ from base to reduplication is defined as
\[
\bm{s}_r = \bm{v}_r - \bm{v}_b.
\]
The shift vector $\bm{s}_r$ captures both the direction and the extent of semantic change from the base words to the corresponding reduplicative constructions. For example, for the reduplication 健健康康 \textit{jian4-jian4-kang1-kang1} `healthy and well' and its base word 健康 \textit{jian4-kang1} `healthy', the shift vector is
\[
\bm{s}_{\text{健健康康}}=
\overrightarrow{\text{健健康康}}
-
\overrightarrow{\text{健康}}.
\]
Here, $\bm{s}_{\text{健健康康}}$ captures the semantic change from the base word 健康 \textit{jian4-kang1} to the reduplicated form 健健康康 \textit{jian4-jian4-kang1-kang1}. 

\subsection{Semantic transparency: base word similarity in shift space}

\begin{figure}[htbp]
  \centering
  \includegraphics[width=0.7\linewidth]{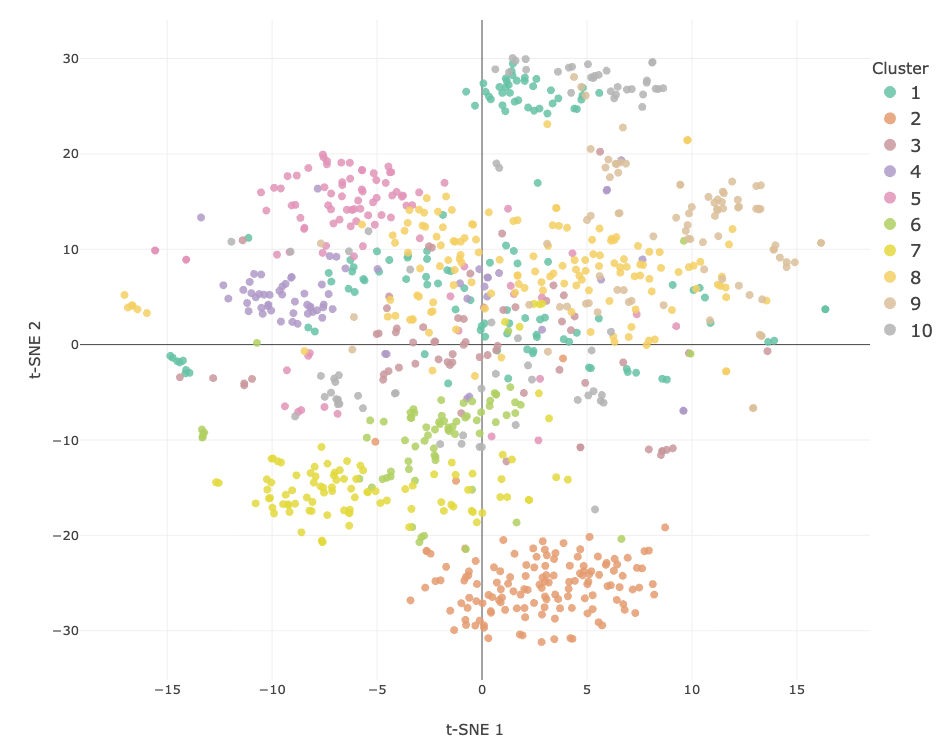}
  \caption{Two-dimensional t-SNE map of shift vectors from base words to reduplicative constructions, colored by the clusters of reduplicative constructions identified by the k-means algorithm.}
  \label{fig:shift_space_colored_original_clusters}
\end{figure}

Focusing on the reduplicative constructions for which a base word exists (955 out of the total 1,010 reduplications), we projected the shift vectors onto a two-dimensional t-SNE space. The resulting t-SNE map is shown in Figure~\ref{fig:shift_space_colored_original_clusters}. Data points are colored according to the clusters identified by the k-means clustering (see section~\ref{sec:Clustering}). This makes it possible to ask to what extent the original clusters of reduplications are also visible in the shift space. If the original clusters remain reasonably well separated, this suggests that the semantic change from base to reduplication is relatively consistent within clusters. If, by contrast, the clusters become more intermixed, this points to more variable or less transparent semantic transformations from base word to reduplication. 

Compared with the semantic space of the reduplications themselves, shown in the left panel of Figure~\ref{fig:tsne_cluster}, it is clear that considerable  structure remains visible in the shift space, although not equally clearly for all clusters. Cluster 2 (orange, bottom center) remains relatively compact. This cluster is dominated by verbal reduplicative constructions with strong eventive semantics, and the corresponding semantic shifts serve to shorten, soften, and render more tentative the event meaning of the base word. For example, the base word 琢磨 \textit{zuo2-mo2} means `to figure out', whereas the reduplicated form 琢磨琢磨 \textit{zuo2-mo0-zuo2-mo0},`think it over a bit', expresses a lighter and more tentative version of the same process. Similarly,  讨论 \textit{tao3-lun4} means `to discuss', while 讨论讨论 \textit{tao3-lun4-tao3-lun4} means `have a brief discussion'.  Words that are used across different word categories, such as 热闹 \textit{re4-nao0}, which can be interpreted as an adjective (`lively'), a verb (`liven up'), or a noun (`scene of bustle'), appear in cluster 2 with the ABAB form 热闹热闹 \textit{re4-nao0-re4-nao0}, `to liven things up a bit'. By contrast, the AABB form 热热闹闹 \textit{re4-re4-nao0-nao0}, `very lively', is associated with the adjectival interpretation of the base word, and belongs to Cluster 8.

Clusters 4 (purple) and 5 (pink), in the outer upper left quadrant, are also relatively well preserved in the shift space. The formations in Cluster 4 change the meanings of their base words by increasing their sensory perceptibility. This can be seen in 高高大大 \textit{gao1-gao1-da4-da4} `burly and imposing', cf. 高大 \textit{gao1-da4} `tall and big', where reduplication makes the size more visually salient. A similar effect is found in the tactile domain. Both 冰凉冰凉 \textit{bing1-liang2-bing1-liang2} `gelid' and 冰冰凉凉 \textit{bing1-bing1-liang2-liang2} `(pleasantly) ice cool', cf. 冰凉 \textit{bing1-liang2} `ice-cold', construe the base property as a more immediate bodily sensation. The same tendency is also visible in the domain of color, as in 通红通红 \textit{tong1-hong2-tong1-hong2} `red through and through', cf. 通红 \textit{tong1-hong2} `bright red'. 

The shift in Cluster 5 involves emphatic reinforcement of the meaning of the base word. This can be seen in 扎扎实实 \textit{zha1-zha1-shi2-shi2} `in a down-to-earth manner', cf. 扎实 \textit{zha1-shi2} `solid', where reduplication reinforces the sense of firmness and reliability. Similarly, 清清白白 \textit{qing1-qing1-bai2-bai2} `completely blameless', cf. 清白 \textit{qing1-bai2} `innocent', presents the state as complete and unequivocal, rather than simply stronger in degree. In 快快乐乐 \textit{kuai4-kuai4-le4-le4} `happily; happy and carefree', cf. 快乐 \textit{kuai4-le4} `happy; happiness', the reduplicated form foregrounds a sustained affective state. 

This semantic shift is often accompanied by a stronger tendency toward adverbial use, as reflected in the occurrence before the adverbial particle 地 \textit{de}, where the reduplicated form modifies a following verb phrase, as in 清清白白地生活 \textit{qing1-qing1-bai2-bai2 de sheng1-huo2-zhe} `live blamelessly', 扎扎实实地工作 \textit{zha1-zha1-shi2-shi2 de gong1-zuo4} `do a solid job', 快快乐乐地上学 \textit{kuai4-kuai4-le4-le4 de shang4-xue2} `go to school happily'. The proximity of Clusters 4 and 5 in the shift space reflects that both clusters involve reinforcement of the base meaning: the reduplicative shift in Cluster 4 emphasizes the intensity of the sensory perception expressed by the base word, whereas the shift in Cluster 5 emphasizes the desirability or reliability of the --- typically positive --- meaning of the base word. 

At the same time, some clusters, such as Cluster 3 (brown), Cluster 8 (light yellow), and Cluster 9 (dark yellow), become more diffuse in the shift space than in the original reduplication space. Among them, Cluster 3 shows the greatest dispersion, spreading across all four quadrants of the shift space.

In the first quadrant, the reduplicative constructions of Cluster 3 are clustered around the repeated actions encoded in the base words, like 颠颠簸簸 \textit{dian1-dian1-bo3-bo3} `bump along the way' cf. 颠簸\textit{dian1-bo3} `bump', 沸沸腾腾\textit{fei4-fei4-teng2-teng2},`bubbling and seething', cf. 沸腾 \textit{fei4-teng2} `boil'. 分分合合 \textit{fen1-fen1-he2-he2} `repeated separations and reunions', cf.分合\textit{fen1-he2} `separation and reunion'. 

By contrast, the reduplicative constructions of Cluster 3 in the fourth quadrant are organized around the quantification of the entities denoted by the base words, such as 家家户户 \textit{jia1-jia1-hu4-hu4} `every family', cf. 家户 \textit{jia1-hu4} `household', 里里外外 \textit{li3-li3-wai4-wai4} `all aspects; completely', cf. 里外 \textit{li3-wai4} `inside and outside' and 老老少少 \textit{lao3-lao3-shao4-shao4} `people of all ages', cf. 老少 \textit{lao3-shao4} `old and young'. 

In the second quadrant, the reduplicative constructions of Cluster 3 are organized around vertical iteration, as exemplified by 沉沉浮浮 \textit{chen2-chen2-fu2-fu2} `sinking and floating repeatedly, floating life', cf. 沉浮 \textit{chen2-fu2} `sink and float', 起起伏伏 \textit{qi3-qi3-fu2-fu2} `repeated ups and downs', cf. 起伏 \textit{qi3-fu2} `up and down' and 坎坎坷坷 \textit{kan3-kan3-ke3-ke3} `repeatedly bumpy', cf. 坎坷 \textit{kan3-ke3} `bumpy'.

The reduplicative constructions of Cluster 3 in the third quadrant comprise formations such as 进进出出 \textit{jin4-jin4-chu1-chu1} `in and out frequently' (cf. 进出 \textit{jin4-chu1} `in and out') and 来来去去 \textit{lai2-lai2-qu4-qu4} `comes and goes repeatedly' (cf. 来去 \textit{lai2-qu4} `come and go'), both of which express repeated coming and going, with the former more prevalent with human subjects (68.9\% according to CCL database), as in 孩子们进进出出 \textit{hai2-zi-men jin4-jin4-chu1-chu1} `the children run in and out', and the latter more readily extended to abstract subjects (35.3\%\footnote{The examples and percentage were retrieved and calculated from the Chinese--English bilingual section of the CCL Corpus Search System}) and more global situations, such as the coming and going of generations or the ebb and flow of people between countries, as in 苦与乐来来去去 \textit{ku3 yu3 le4 lai2-lai2-qu4-qu4} `suffering and happiness come and go'.

Formations such as 前前后后 \textit{qian2-qian2-hou4-hou4} `throughout, completely' (cf. 前后 \textit{qian2-hou4} `front and back') express a vivid sense of step-by-step thoroughness and comprehensiveness, similar to the formations in the fourth quadrant. For instance, 前前后后、里里外外叙述个遍 \textit{qian2-qian2-hou4-hou4, li3-li3-wai4-wai4 xu4-shu4 ge bian4} literally translates as `front-front-back-back, inside-inside-outside-outside narrate CLASSIFIER once-through', meaning `Tell me, covering every detail, inside and out'.

Other clusters overlap more strongly with neighboring regions than they did in the original reduplication space. 
For example, Cluster 1 and Cluster 10 appear to overlap more strongly in the shift space. Items in both clusters share the same constructional pattern, 一 \textit{yi1} `one' + classifier + 一 \textit{yi1} `one' + classifier, which expresses a unit-by-unit meaning. Examples include 一年一年 \textit{yi4-nian2-yi4-nian2} `year after year', cf. 一年 \textit{yi4-nian2} `one year', and 一行一行 \textit{yi4-hang2-yi4-hang2} `line by line', cf. 一行 \textit{yi4-hang2} `one line', in Cluster 10; 
and 一片一片 \textit{yi2-pian4-yi2-pian4} `piece by piece', cf. 一片 \textit{yi2-pian4} `one piece', and 一桶一桶 \textit{yi4-tong3-yi4-tong3} `bucket by bucket', cf. 一桶 \textit{yi4-tong3} `one bucket', in Cluster 1.
 
The comparison between the reduplication space and the shift space shows that some clusters remain relatively compact in both spaces, suggesting that these reduplicative constructions are similar not only in the meanings of the reduplicative forms themselves, but also in the semantic shifts from base word to reduplicative form (e.g., Clusters 2, 4, and 5). Other clusters become more dispersed in the shift space than in the reduplication space, indicating that a single cluster may involve several different types of shift change from base to reduplication (e.g., Clusters 3, 8, and 9). Conversely, some clusters that are relatively distant in the reduplication space become closer or more overlapping in the shift space, suggesting that they may instantiate similar types of semantic change despite belonging to different clusters in the reduplication space (e.g., Clusters 1 and 10).

To examine the structure of the shift space in more detail, the next section investigates what types of semantic shift from base words to reduplicative forms can be identified.

\subsection{Types of base-to-reduplication shifts}

The preceding analysis showed that some reduplication clusters remain relatively compact in the shift space, whereas others become more diffuse or overlap strongly with neighboring regions, suggesting that the semantic shifts from base words to reduplicative constructions may themselves be structured into several types. In this section, we therefore examine the structure of the shift space in order to identify shift types, and further ask whether these shift types are associated differently with the two reduplicative patterns, AABB and ABAB.

To identify shift types, we applied k-means clustering to the 955 shift vectors for reduplicative constructions whose corresponding base words exist.  Different values of $k$ in the range [2, 15] yielded very similar results, with LDA cross-validation accuracies ranging from 0.97 for $k=2$ to 0.84 for $k=14$. For our exploration, we opted for $k=10$, the number of clusters previously observed for the semantic space of the reduplications. Under leave-one-out cross-validation, LDA classification accuracy for these 10 clusters reached 87.33\%, substantially above the majority-class baseline of 16.65\%.  

\begin{figure}[H]
  \centering
  \includegraphics[width=\textwidth]{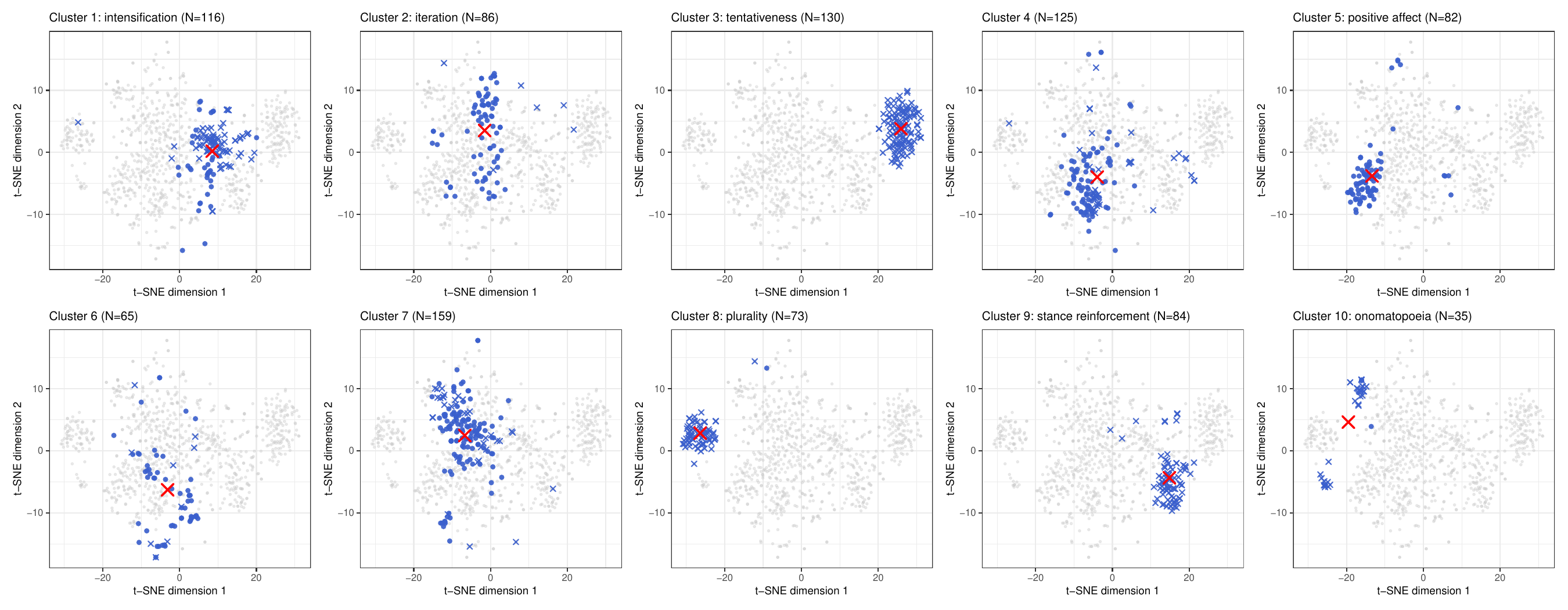}
  \caption{Two-dimensional t-SNE map of shift vectors for Mandarin reduplicative constructions. Each panel highlights one k-means cluster in the shift-vector space, with the remaining data points shown in grey. Circle-shaped and cross-shaped data points indicate AABB and ABAB forms respectively. Red crosses denote cluster centroids.}
  \label{fig:shift_clusters}
\end{figure}

For visualization, the shift vectors were projected from the shift space onto a two-dimensional t-SNE map, shown in Figure~\ref{fig:shift_clusters}. Each panel highlights one cluster in blue, with the remaining data points shown in grey. Circle-shaped and cross-shaped data points represent AABB and ABAB constructions respectively. Red crosses mark cluster centroids.  Most clusters are reasonably well separated, but cluster 6 is of low quality and overlaps largely with cluster 7. On the other hand, some clusters (e.g., 3, 5, 8, 9, and 10) are clearly distinct in the t-SNE projection.  Furthermore, some clusters have a much wider scatter than others (compare, e.g., 1 and 2 with 3 and 8).  Although the clustering is not perfect, it is useful as a guide to the shift space.

Figure~\ref{fig:shift_clusters} shows that some shift clusters are dominated by ABAB, such as Clusters 3, 8, and 9, and that others are dominated by AABB, such as Cluster 5. A majority of the clusters contain both patterns (Clusters 1, 2, 4, 6, 7, 10).  Nevertheless, LDA leave-one-out classification accuracy for AABB and ABAB reached 92.88\%, far above the majority-class baseline (50.89\%). This high classification accuracy is consistent with the hypothesis that the two constructions realize different types of semantic shifts.  The high classification accuracy differs from the clusters in the t-SNE map.  This is likely due to the t-SNE that necessarily highlights the major `latent' dimensions of variation in the shift space, whereas the LDA can take into account the full information in this space.

The semantic shifts within the individual clusters differ in their degree of coherence. Some clusters, such as Clusters 1, 3, and 5, have similar semantic shifts among their members and therefore receive descriptive labels in Figure~\ref{fig:shift_clusters}. Other clusters contain members with several semantic shifts and therefore do not receive a single cluster-level label. For these clusters, we will describe their internal composition and the types of the semantic shifts.

Cluster 5, labeled \textit{positive affect}, consists entirely of AABB forms (n=82), and the semantic shifts in this cluster tend to construe the meaning of the base words in a more positive way. For instance, 安静 \textit{an1-jing4} `quiet' is relatively neutral as a base word, denoting the absence of noise, as in the imperative 安静! \textit{an1-jing4} `Be quiet!' or in predicative uses such as 那地方很安静 \textit{na4 di4-fang hen3 an1-jing4} `That place is quiet'. By contrast, 安安静静 \textit{an1-an1-jing4-jing4} profiles quietness as something maintained over the course of an event, often with the implication that this is the appropriate or expected way. In expressions such as 安安静静地睡觉 \textit{an1-an1-jing4-jing4 de shui4-jiao4} `sleep peacefully', the reduplication expresses a clearer evaluative stance of the speaker than the base word 安静 \textit{an1-jing4} `quiet'. A similar tendency can also be observed with base words expressing negative affect (n=6). For instance, the base word 孤单 \textit{gu1-dan1} means `lonely', as in 孤单的老人 `a lonely elderly person', and straightforwardly denotes a specific negative emotional state. By contrast, 孤孤单单 \textit{gu1-gu1-dan1-dan1} means `alone', as in 留下我一个人孤孤单单的 \textit{liu2-xia4 wo3 yi1-ge4 ren2 gu1-gu1-dan1-dan1 de} `leaving me alone'.

When the base word is already positive, reduplication reinforces this positive evaluation and often shifts the reduplicative constructions from the adjective toward adverbial use. This can be seen in the contrast between 开心 \textit{kai1-xin1} `happy' and 开开心心 \textit{kai1-kai1-xin1-xin1} `happily'. Concordance results \footnote{The concordance check was based on the first 1,000 lines inspected for each form in the CCL corpus \url{https://corpus.pku.edu.cn/}, after restricting the search to texts from the 2000s to 2020s.} show that the most frequent immediate left collocate of 开心 \textit{kai1-xin1} `happy' is 很 \textit{hen3} `very' (12\%), like in 很开心 \textit{hen3 kai1-xin1} `very happy', while the most frequent immediate right collocate is 的 \textit{de} (13\%), an attributive marker used in nominal modification, like 开心的回忆 \textit{kai1-xin1 de hui2-yi4} `happy memory'. By contrast, 开开心心 is followed by the adverbial marker 地 \textit{de} significantly more often than 开心 (25.0\% vs. 9.0\%, $p < 0.01$, proportions test), as in 开开心心地过大年 \textit{kai1-kai1-xin1-xin1 de guo4 da4-nian2} `celebrate the New Year happily'. It is also followed by the event verb 过 \textit{guo4} `spend (time)' more often than 开心 (10.0\% vs. 0.0\%, $p < 0.01$, proportions test), further suggesting a stronger preference for event-modifying, adverbial use.

Cluster 1 contains both AABB (n=38) and ABAB (n=78) forms, with the shift vectors amplifying the base meaning along scalar dimensions such as degree, quantity, and spatial coverage. A small number of AABB and ABAB forms (n=5) in this cluster also share the same base word.
Among the AABB forms, this tendency can be illustrated by 上下 \textit{shang4-xia4} `up and down' and 上上下下 \textit{shang4-shang4-xia4-xia4} `from top to bottom; throughout'. Whereas 上下 \textit{shang4-xia4} encodes a basic vertical contrast, 上上下下 \textit{shang4-shang4-xia4-xia4} extends this contrast into more exhaustive spatial coverage. 
According to the concordance results, 上上下下 \textit{shang4-shang4-xia4-xia4} is followed by 都 \textit{dou1} `all' more often than 上下 \textit{shang4-xia4} (18.0\% vs. 1.0\%, $p < 0.001$, proportions test), as in 上上下下都非常团结 \textit{shang4-shang4-xia4-xia4 dou1 fei1-chang2 tuan2-jie2} `people at all levels are very united'. By comparison, 上下 \textit{shang4-xia4} `up and down' is followed by 都 \textit{dou1} `all', in only 1.8\%, and these cases typically occur with an overt collective expression headed by 全 \textit{quan2} `whole', as in 全国上下都 \textit{quan2-guo2 shang4-xia4 dou1} `the whole country all'. 

The base forms of the ABAB members in this cluster often begin with degree modifiers such as 太 \textit{tai4} `too', 很 \textit{hen3} `very', and 好 \textit{hao3} `quite', as in 太多 \textit{tai4-duo1} `too many', 很冷 \textit{hen3-leng3} `very cold', and 好远 \textit{hao3-yuan3} `quite far'. Across the \textit{tai4}-forms in this cluster, the intensifier 实在 \textit{shi2-zai4} `indeed' occurs before the reduplicated forms more often than before the corresponding base forms (4.9\% vs. 3.1\%, $p < 0.05$, proportions test), suggesting a shift toward a more emphatic scalar construal.

Some base words in this cluster give rise to both reduplicated patterns. For example, 许多 \textit{xu3-duo1} `many' gives rise to both the AABB form 许许多多 \textit{xu3-xu3-duo1-duo1} and the ABAB form 许多许多 \textit{xu3-duo1-xu3-duo1}. The base form already encodes plurality, but the reduplicated forms strengthen this meaning. This is reflected in the fact that 许多许多 is preceded by 还有 \textit{hai2-you3} `there are still' more often than 许多 (13.2\% vs. 2.4\%, $p < 0.001$, proportions test), while 许许多多 shows the same tendency, though more weakly (4.5\% vs. 2.4\%, $p < 0.05$, proportions test). The reduplicated forms also occur more often with nouns such as 故事 \textit{gu4-shi4} `story' and 事例 \textit{shi4-li4} `case' in the immediate right context: 2.0\% for 许多许多 and 1.3\% for 许许多多, compared with 0.2\% for 许多 ($p < 0.001$ and $p < 0.01$, respectively, proportions test), yielding readings such as `many, many stories' and `numerous cases'. A similar change can be observed for 永远 \textit{yong3-yuan3} `forever'. The base form already encodes temporal persistence, while the reduplicated form 永远永远 \textit{yong3-yuan3-yong3-yuan3} reinforces this temporal meaning. In the concordance sample, 永远永远 is followed by 爱 \textit{ai4} `love' much more often than 永远 (7.1\% vs. 0.2\%, $p < 0.001$, proportions test), suggesting that reduplication extends the temporal meaning of the base toward a stronger construal of enduring commitment; cf. also 永远永远忘不了 \textit{yong3-yuan3-yong3-yuan3 wang4 bu4 liao3} `never, ever forget'.

Cluster 3 is highly homogeneous in formal terms, consisting entirely of ABAB forms (n = 130). The semantic shifts in this cluster tend to recast the base event as more delimited, more tentative, and more interactionally softened. This can be seen in the contrast between 琢磨 \textit{zuo2-mo2} `figure out; ponder' and 琢磨琢磨 \textit{zuo2-mo0-zuo2-mo0} `think it over a bit'. Whereas the base verb 琢磨 \textit{zuo2-mo2} can denote sustained or serious mental engagement, the reduplicated form more often presents the event as a bounded episode of consideration. Concordance results show that 再 \textit{zai4} `again; further' occurs immediately before 琢磨琢磨 significantly more often than before 琢磨 (7.0\% vs. 0.4\%, $p < 0.001$, proportions test), as in 再琢磨琢磨 \textit{zai4 zuo2-mo0-zuo2-mo0} `think it over a bit more'. Similarly, 好好 \textit{hao3-hao3} `carefully' also occurs immediately before 琢磨琢磨 more often than before 琢磨 (11.1\% vs. 1.2\%, $p < 0.001$, proportions test), as in 好好琢磨琢磨 \textit{hao3-hao3 zuo2-mo0-zuo2-mo0} `think it over carefully'. Together, these patterns suggest that the reduplicated form is more readily used in contexts of bounded, exploratory reconsideration.

Some clusters do not support a single cluster-level label for their shift vectors (Clusters 4, 6, and 7). Cluster 6, for example, is internally differentiated and contains several subtypes of semantic shifts. One subtype is configurational depiction, as in 高低 \textit{gao1-di1} `high and low' and 高高低低 \textit{gao1-gao1-di1-di1} `uneven'. The base form 高低 \textit{gao1-di1} `high and low' expresses a simple contrast in height, and as a noun means `height'. By contrast, 高高低低 \textit{gao1-gao1-di1-di1} `uneven' presents contrasts in height as  distributed in space. The concordance results show that 高高低低 is followed by 的 \textit{de}, the attributive marker, in 46.2\% of instances, compared to 6.7\% for 高低, indicating a strong preference for use as an adjective. 高高低低 also occurs more often in descriptions of scenes, with words such as 路 \textit{lu4} `road', 山 \textit{shan1} `mountain', 树 \textit{shu4} `tree', 屋 \textit{wu1} `house', and 建筑 \textit{jian4-zhu4} `building' (50.2\% vs. 20.8\%, $p < 0.01$, proportions test). 
A second subtype involves onomatopoeia (note, however, that a majority of onomatopoeia are found in cluster 10). For instance, the base word 叮咚 \textit{ding1-dong1} `tinkle' has as reduplication 叮叮咚咚 \textit{ding1-ding1-dong1-dong1} `tinkling repeatedly', which typically occurs in rhythmic contexts with words such as 音乐 \textit{yin1-yue4} `music', 节奏 \textit{jie2-zou4} `rhythm', 琴 \textit{qin2} `instrument', 鼓点 \textit{gu3-dian3} `drumbeat', and 乐音 \textit{yue4-yin1} `musical sound' (21.5\% vs. 12.4\%, $p < 0.01$, proportions test). The semantic shift recasts a sound label as the depiction of a repeated sound.
The last subtype shifts the base event toward a habitual construal. The base form 唱跳 \textit{chang4-tiao4} `sing and dance' simply denotes a combination of two  activities, whereas 唱唱跳跳 \textit{chang4-chang4-tiao4-tiao4} `singing and dancing' presents this activity as a repeated habitual event. In the inspected CCL concordance samples, compared to 唱跳, 唱唱跳跳 occurs more often in texts discussing cultural activities, cf. 文艺活动无非是唱唱跳跳,玩玩乐乐 \textit{wen2-yi4 huo2-dong4 wu2-fei1 shi4 chang4-chang4-tiao4-tiao4, wan2-wan2-le4-le4 de} `literary and artistic activities are nothing more than singing, dancing and having fun', and 过年过节唱唱跳跳 \textit{guo4-nian2-guo4-jie2 chang4-chang4-tiao4-tiao4} `singing and dancing during the holidays and festivals (14.9\% vs. 5.7\%, $p < 0.01$, proportions test). 唱唱跳跳 is also more often associated with temporal expressions such as 整天 \textit{zheng3-tian1} `all day' and 成天 \textit{cheng2-tian1} `all day long' (8.1\% vs. 0.6\%, $p < 0.01$, proportions test).
The analysis of shift-vectors  clarifies how base words are related to their corresponding reduplicative forms. However, for 5.4\% of the reduplicative constructions, no corresponding base word was included in the analysis. The next section therefore turns from individual base-reduplication pairs to the comparison of the structures of the base-word space and the reduplication space by means of Procrustes analysis.

\section{Procrustes analysis}
\label{sec:procrustes}

The analyses presented thus far have shown that reduplicative constructions show considerable clustering in both the semantic space and the shift space. The calculation of shift vectors requires that there is a base word that can be matched to the reduplication. However, for some 5.4\% of the reduplications, no base word exists.  To include these words in our analyses, we made use of Procrustes analysis.

Procrustes analysis is a statistical method originally developed for comparing shapes, such as the shapes of leaves. We use it to compare the embeddings of base words with the embeddings of the reduplications.   The basic idea is that if sets of  points in a high-dimensional space have a similar structure, then after differences in location, scale, and orientation are removed, a rotation should be sufficient to bring one set of points into close correspondence with the other set of points. In linguistic studies, Procrustes analysis has been used to compare semantic spaces across languages \citep{YangBaayen2025, Mohiuddin2020}, and to align word representations across different developmental stages \citep{JorgeBotana2018}. In the present study, we use procrustes analysis to examine whether the semantic space of base words and the semantic space of reduplicative constructions can be brought into close correspondence.  If the reduplication space can be aligned with the base-word space with only minor distortion, this will indicate that reduplications largely preserve the semantic organization of their  base words.

We calculated the Procrustes alignment based on the centroids of the 10 clusters given by the k-means algorithm for the base words, and the centroids of the 10 clusters that we obtained for the reduplications, also using the k-means algorithm.
However, because the two spaces were clustered independently,  their numerical cluster labels are not directly comparable. For example, cluster 1 in the base-word space does not necessarily correspond to cluster 1 in the reduplication space. We therefore computed the pairwise cosine similarities between all base-word cluster centroids and all reduplication cluster centroids, and used the Hungarian algorithm \citep{Kuhn1955} to select the one-to-one matching that maximized the total similarity between paired centroids:
\[
\max_{\pi} \sum_{i=1}^{K} \cos\left(\mathbf{b}_i, \mathbf{r}_{\pi(i)}\right)
\]
where $\mathbf{b}_i$ is the centroid of cluster $i$ in the base-word space, $\mathbf{r}_j$ is the centroid of cluster $j$ in the reduplication space, and $\pi$ is a permutation of cluster indices. The Hungarian algorithm therefore finds the one-to-one assignment of reduplication clusters to base-word clusters that maximizes the overall cosine similarity between matched centroid pairs. Operationally, because the Hungarian algorithm solves a cost-minimization problem, cosine similarities were converted into costs by subtracting each similarity value from the maximum similarity in the matrix. Minimizing this cost is equivalent to maximizing the total cosine similarity between matched centroids. 

An asymmetric Procrustes alignment was then carried out on the resulting paired cluster centroids. We used an asymmetric alignment because our goal was to inspect the reduplication space in relation to the base-word space. The base-word centroid configuration was treated as the reference configuration, and the matched reduplication centroid configuration was transformed to best fit it. The Procrustes transformation was therefore estimated from 10 matched pairs of 200-dimensional centroid vectors, and was subsequently applied to all reduplication vectors to obtain Procrustes-rotated reduplication vectors in the base-word semantic space.

The analysis was carried out using the \texttt{procrustes} and \texttt{protest} functions from the \texttt{vegan} package \citep{Oksanen2022}, with significance assessed by 999 permutations. The Procrustes analysis yielded a high correlation between the two centroid configurations, \(r = 0.9336\), with a residual sum of squares of \(m_{12}^{2} = 0.1284\). The fit was significant under permutation testing, \(p = 0.001\), indicating that the base-word and reduplication centroid configurations share a highly similar global geometry.

\begin{figure}[H]
    \centering
    \begin{subfigure}{0.32\linewidth}
        \centering
        \includegraphics[width=\linewidth]{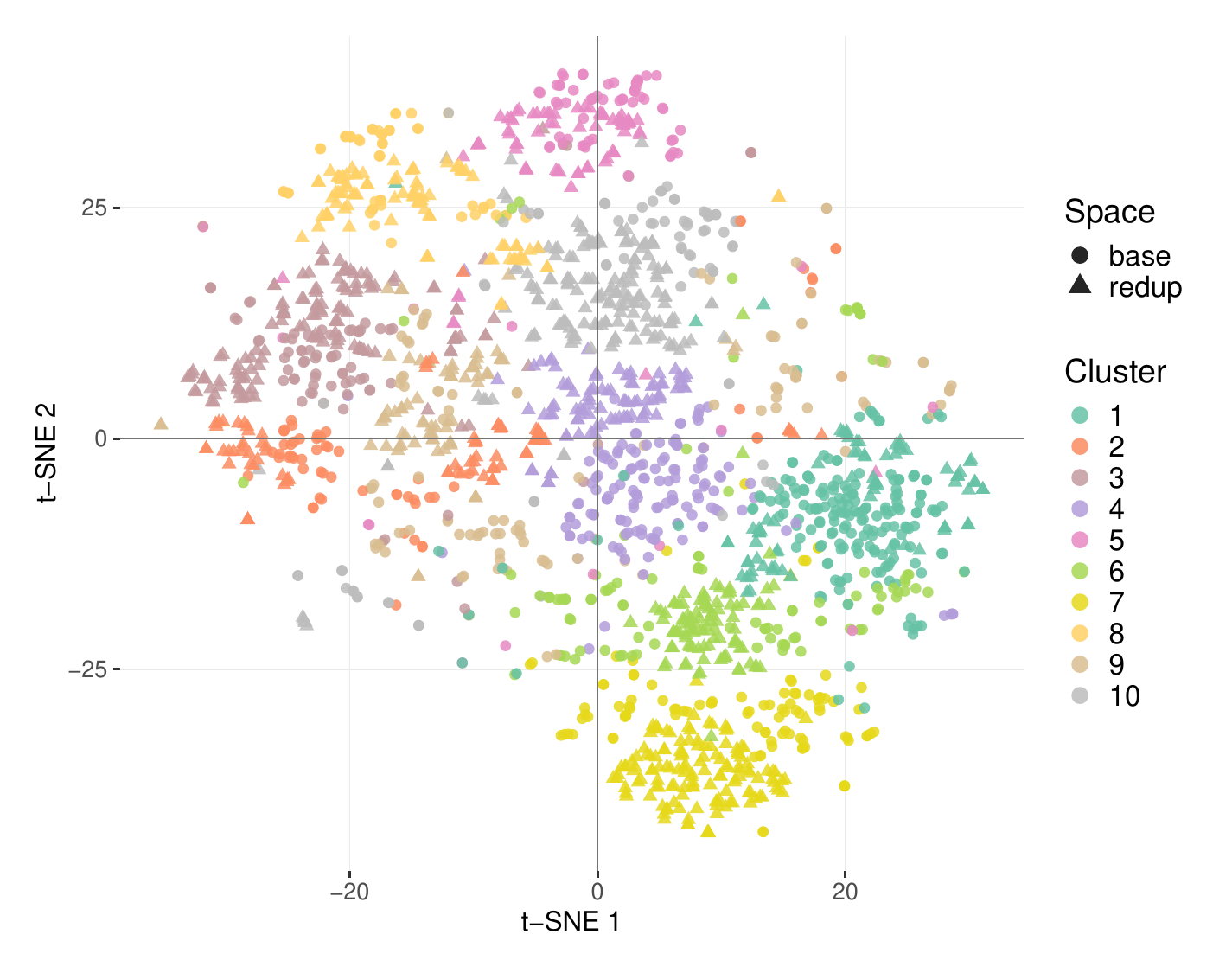}
        \label{fig:base_redup_procrustes_tsne}
    \end{subfigure}
    \hfill
    \begin{subfigure}{0.32\linewidth}
        \centering
        \includegraphics[width=\linewidth]{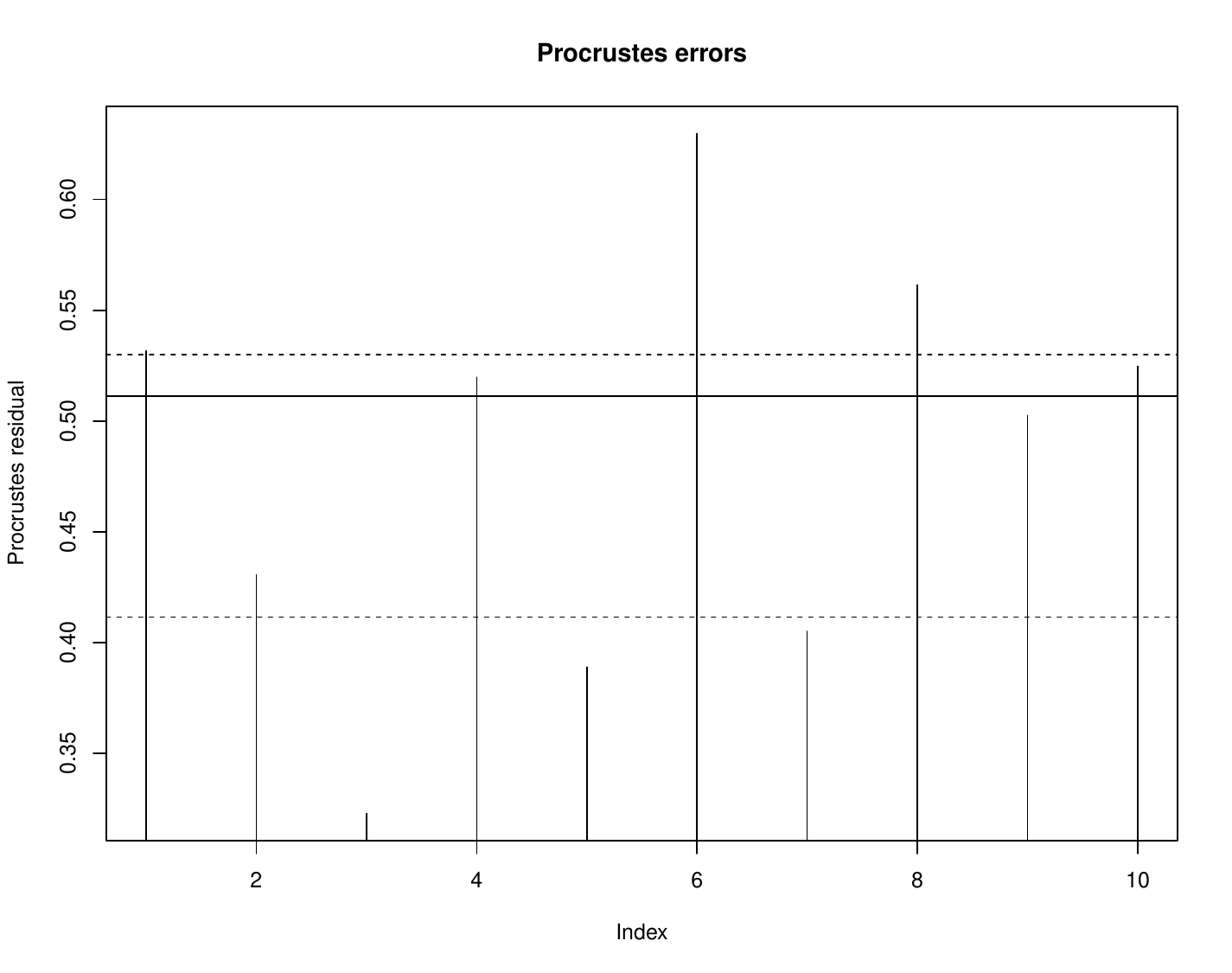}
        \label{fig:base_redup_procrustes_errors}
    \end{subfigure}
    \hfill
    \begin{subfigure}{0.29\linewidth}
        \centering
        \includegraphics[width=\linewidth]{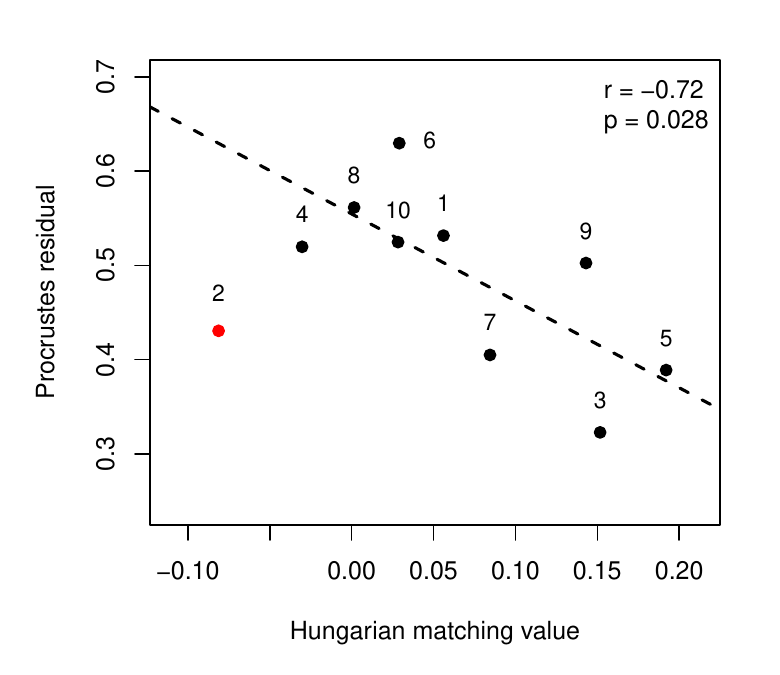}
        \label{fig:vh_procrustes_residuals}
    \end{subfigure}
    \caption{Procrustes analysis of the base word and reduplicative construction spaces. Left panel: two-dimensional t-SNE map of base-word vectors and Procrustes-rotated reduplication vectors. Circles indicate base words and triangles indicate reduplicative constructions. Colors indicate the cluster correspondences by Hungarian matching. Middle panel: cluster-level Procrustes residuals for the matched centroid pairs. Each vertical line represents the residual for one matched base-reduplication cluster pair after Procrustes alignment. Larger residuals indicate clusters for which the rotated reduplication centroid remains farther away from the corresponding base-word centroid, like Clusters 6, 8, 1, 10 and 4. Right panel: regression plot relating Hungarian matching values to Procrustes residuals. Each point represents one matched cluster after Hungarian matching, with the number indicating the cluster label. Cluster 2 (Cook's distance $>2$) was excluded from this regression plot. The negative association indicates that clusters with higher Hungarian matching values tend to have smaller Procrustes residuals after alignment ($r = -0.72$, $p = .028$).}
    \label{fig:procrustes_BR}
\end{figure}

The resulting Procrustes transformation was then applied to all reduplication vectors, yielding Procrustes-transformed reduplication vectors in the base-word semantic space. These aligned reduplication vectors were visualized together with the base-word vectors using t-SNE.  The left panel of Figure~\ref{fig:procrustes_BR} shows the aligned semantic space. Circles indicate base words, triangles indicate reduplicative constructions, and colors indicate the cluster correspondences established by Hungarian matching. Clusters in which base-word and reduplication points overlap closely provide evidence for high-quality local alignment, whereas clusters that remain more clearly separated indicate greater divergence between the two spaces. 

The middle panel presents the cluster-level Procrustes residuals for the matched centroid pairs. These residuals quantify how far each rotated reduplication centroid remains from its corresponding base-word centroid after alignment, with the smaller residuals indicating the higher degree of semantic continuity
(or semantic transparency) between the base words and the reduplicative constructions, like clusters 3, 5, 7, 2 and 9, while larger residuals indicating stronger semantic reorganization, like clusters 6, 8, 1, 10 and 4.

As shown in the middle panel of Figure~\ref{fig:procrustes_BR}, Cluster 3 has the smallest Procrustes residual among the ten matched cluster pairs (\(0.323\)), whereas Cluster 6 has the largest residual (\(0.630\)). A smaller residual indicates that the rotated reduplication centroid can be brought into closer correspondence with its matched base-word centroid, while a larger residual indicates that a greater discrepancy remains after alignment. The contrast between clusters 3 and 6 is consistent with the results of the Hungarian matching: The matched centroid pair for Cluster 3 has a cosine similarity of \(0.87\), which is higher than the corresponding value of \(0.80\) for Cluster 6.  

In fact, the Procrustes residuals are negatively correlated with the Hungarian correlation score ($r = -0.72, t(7) = -2.766, p = 0.028$, after removal of one outlier (Cook's distance $>2$), cluster 2, the datapoint with the largest negative Hungarian correlation score), as shown in the right panel of Figure~\ref{fig:procrustes_BR}.  Clearly, the quality of the Procrustes analysis depends on the quality of the Hungarian alignment of the base word clusters and the reduplication clusters.

The magnitude of the residuals also corresponds with the distribution of the clusters in the left panel of Figure~\ref{fig:procrustes_BR}. In Cluster 3, the brown points in the upper-left region, the circles and triangles are well mixed, suggesting that the base-words and rotated reduplications occupy highly similar regions of the semantic space. By contrast, the light green points of Cluster 6 are more dispersed. The triangles are concentrated more centrally, while many of the circles are distributed to the left or to the right, suggesting that the base-word and rotated reduplication vectors do not align even after Procrustes rotation.

Both Hungarian matching correlations and the Procrustes residuals can be seen as measures of semantic transparency, as they gauge the extent to which the meaning of a reduplicative construction remains predictable from its base word. A high transparency is expected to be reflected not only in the Hungarian correlation and in the Procrustes residual, but also in a greater degree of mixing between the base-word and rotated reduplication vectors in the semantic space, as visualized in the left panel of Figure~\ref{fig:procrustes_BR}.

Cluster 3 (represented by the dark-brown points in the upper-left region) shows the highest degree of semantic transparency. In this cluster, the reduplicative construction preserves the core semantic features of the base word, while elaborating one of its inherent semantic dimensions through nominal plurality, event iteration and scalar intensification.
For example, the base noun compound 枝叶\textit{zhi1-ye4} collectively refers to `branches and leaves', whereas 枝枝叶叶\textit{zhi1-zhi1-ye4-ye4} `many branches and leaves' expresses greater plurality of the same entities. These two words show a high degree of semantic transparency, with a cosine similarity of 0.781, because both refer to branches and leaves and just differ in referential plurality. As for the event iteration, the base word 明灭 \textit{ming2-mie4} denotes an action of `light on and off', whereas 明明灭灭 \textit{ming2-ming2-mie4-mie4} `keep flickering on and off' foregrounds this action as more recurrent. Their cosine similarity of 0.866 indicates that reduplication preserves the core action meaning while amplifying its event iteration. The base word 模糊 \textit{mo2-hu2} `vague'  and its reduplicative construction 模模糊糊 \textit{mo2-mo2-hu2-hu2} `very vague' show a high degree of semantic transparency, with a cosine similarity of 0.782. Both forms denote a lack of perceptual or cognitive clarity, but the reduplication describes the state at a higher degree on the scale of indistinctness, as in 模糊的印象 \textit{mo2-hu2 de yin4-xiang4} `a vague impression' and 模模糊糊的印象 \textit{mo2-mo2-hu2-hu2 de yin4-xiang4} `remarkably vague impression'. 

By contrast, Cluster 6 has the largest Procrustes residual. The reduplicative constructions in this cluster often develop discourse-pragmatic uses that are not fully predictable from the lexical semantics of their bases. For example, the base verb 睡觉 \textit{shui4-jiao4} `sleep' denotes a sleeping event, whereas its ABAB form 睡觉睡觉 \textit{shui4-jiao4-shui4-jiao4} `time to sleep' functions as a directive. The base 走开 \textit{zou3-kai1} `walk away' denotes a motion event, as in 他转身走开了 \textit{ta1 zhuan3-shen1 zou3-kai1 le} `He turned and walked away'. By contrast, 走开走开 \textit{zou3-kai1-zou3-kai1} `go away' is typically used as an independent utterance to perform a directive speech act. 

Cluster 4 (purple dots) contains two  subgroups, one dominated by base words and the other by reduplicative constructions. The cluster has an intermediate degree of cross-space alignment, with a Procrustes residual of 0.520. The reduplicative constructions generally preserve the core lexical meanings of their bases. Many base words in this cluster function as gradable adjectives and can be modified by degree expressions such as 很 \textit{hen3} `very' and 更 \textit{geng4} `more'. However, their reduplicative counterparts generally resist such modification \citep[see, e.g.][]{ZhuJingsong2003, Wang2023}.
For example, 高兴 \textit{gao1-xing4} `happy' can occur in 很高兴 \textit{hen3 gao1-xing4} `very happy' or 更高兴 \textit{geng4 gao1-xing4} `happier', while 高高兴兴 \textit{gao1-gao1-xing4-xing4} normally cannot occur with these degree modifiers. At the same time, the reduplicative construction shows a stronger preference for adverbial functions, as in 高高兴兴地回家 \textit{gao1-gao1-xing4-xing4 de hui2-jia1} `go home cheerfully'. In other words, the separation between base words and reduplications within cluster 4 does not so much reflect a change in lexical meaning, but rather different preferences for use as adjectives or adverbs.

In summary, we used Procrustes rotation to align the semantic spaces of the base words and reduplicative constructions. The overlap between the aligned spaces reflects the general semantic transparency of reduplication, whereas the remaining differences bring its semantic effects. In highly transparent regions, reduplicative constructions largely preserve the core meanings of their bases while amplifying scalar degree, referential plurality, or event iteration. By contrast, the more diffuse regions with larger Procrustes residuals comprise reduplications that are less semantically transparent with respect to their base words,  or that have different pragmatic functions. 

\section{General discussion}
\label{sec:discussion}

The present study reports a quantitative investigation of Mandarin AABB and ABAB reduplicative constructions using distributional semantics.  

Mandarin reduplicative constructions have a range of subtle shades of meaning that have been documented in the literature \citep{chao1968spoken,hua2003reduplication,ZhuJingsong1998, ZhuJingsong2003,Zhang2015}, but that are far from straightforward to characterize precisely. We show that considerable headway can be made for semantic and pragmatic profiling of Mandarin reduplication using embeddings from distributional semantics. 

We first examined whether reduplicative constructions form distinct clusters in semantic space and whether the AABB and ABAB constructions show different semantic distributions. We then profiled these constructions with respect to word category, valence, self-relatedness, and sound-related meaning. For the reduplicative constructions with identifiable base words, we further investigated the semantic shifts of reduplication using shift vectors. Finally, Hungarian matching and Procrustes analysis were used to compare the semantic spaces of the base words and the reduplicative constructions to assess semantic transparency.

A first important result is that several generalizations previously discussed in the descriptive literature on Mandarin reduplication are recovered in the embedding space. AABB and ABAB constructions show systematic semantic differences: they occupy partly different regions of semantic space and have different semantic profiles. AABB constructions are more strongly associated with affective, plural, and sound-related meanings, whereas ABAB constructions are more strongly associated with verbal, interactional, and self-relevant meanings. The two patterns can also be distinguished with high accuracy from both their embeddings and their shift vectors.

This result goes beyond the qualitative descriptions by making the differences between AABB and ABAB more explicit and measurable. Distributional semantics makes it possible to ask not only whether the two patterns differ, but also where in semantic space the differences arise, how strong they are, and which semantic properties contribute to them.

The semantic profiling analyses further clarify that the meaning of a Mandarin reduplicative construction cannot be reduced to a single semantic category. For example, 开开心心 \textit{kai1-kai1-xin1-xin1} `very happy' is strongly positive, relatively adjective- and adverb-like, moderately self-relevant, and only weakly sound-related. This example illustrates that several semantic tendencies may be present in the same construction at the same time.  In fact, the previous literature on Mandarin reduplication offers a bewilderingly wide range of descriptions, including intensification, plurality, iteration, vividness, affect, tentativeness, politeness, and speaker stance. High-dimensional embeddings succeed in capturing all these different facets of meaning jointly. The many descriptions in the literature are not in conflict, but highlight different aspects of the same high-dimensional semantic space. Semantic profiling makes it possible to obtain precise but also linguistically meaningful interpretations of how reduplications structure the  semantic space.

Shift vectors specify the semantic change from a base word to its derived reduplicative construction. Seven types of semantic shifts have been identified in the literature: intensification, iteration, tentativeness, positive affect, plurality, stance reinforcement, and onomatopoeic depiction. These types of semantic shifts are distributed differently across AABB and ABAB constructions. Some types of shift are strongly associated with one particular pattern. For example, tentative meanings were found only among ABAB constructions, and the valence of positive affect is dominant among AABB constructions. 

A Procrustes analysis clarified that the overall organization of the base-word space is largely preserved in the reduplication space. However, this preservation is not equally strong across the semantic space. In some regions, base words and their corresponding reduplicative constructions align closely. Here, reduplication mainly elaborates a semantic property that is already present in the base. For example, 山水 \textit{shan1-shui3} `mountains and waters' and 山山水水 \textit{shan1-shan1-shui3-shui3} `mountains and waters everywhere; landscape' share the same core meaning. Here, the reduplicative construction adds a stronger plural or collective interpretation. In other regions of semantic space, the alignment of base words and reduplications is weaker. For example, 走开 \textit{zou3-kai1} `walk away' denotes a motion event, whereas 走开走开 \textit{zou3-kai1-zou3-kai1} `Go away!' is typically used as a directive. In this example, reduplication introduces additional discourse-pragmatic meanings that are not an intrinsic part of the lexical meaning of the base.

The Procrustes analysis was carried out using the centroids of 10 clusters of base words and 10 clusters of reduplications, which were established independently using the k-means algorithm. We used Hungarian matching to align as best as possible the base word clusters with the reduplication clusters, and then established the Procrustes rotation on the basis of the centroids of these clusters. Both the Hungarian matching and the Procrustes analysis  provide new ways for assessing the transparency quantitatively. Hungarian matching measures how well independently obtained base-word and reduplication clusters correspond to each other, whereas the Procrustes residuals measure how much mismatch remains after alignment.  The two measures are correlated: clusters with better Hungarian matching also tend to have smaller Procrustes residuals. Taken together, the two measures make it possible to identify regions in which reduplicative constructions are more, or less, transparent with respect to their base words.

In the present study, we made use of static embeddings that assign a fixed meaning to each reduplicated word. These static embeddings cannot distinguish between the different meanings and  discourse functions that one and the same form may have in different contexts.  They are `blends' of these different senses and functions that are dominated by those senses and functions that are most frequent. As a consequence, the semantic profiles reported in the present study are approximate, and will benefit considerably by moving from fixed embeddings to contextualized embeddings, the embeddings that can be obtained with large language models applied to word tokens in their discourse context.

In conclusion, distributional semantics makes it possible to see what remains stable and what changes when a Mandarin disyllabic base word is used in the reduplicative construction. It thereby provides a quantitative perspective on the semantic systematicities and the semantic versatility of Mandarin reduplication.  The present study is offered in the hope that the combination of semantic profiling, the investigation of shift vectors, and semantic-space alignment may prove useful for studying other word-formation and construction-formation processes.

\section*{Data availability statement}
The data, analysis code, and interactive figures are available through
an \href{https://osf.io/42p7e/overview?view_only=d026af95c60a4d68a91ea62f03947c52}
{anonymous OSF repository}.

\clearpage
\printbibliography

@book{chao1968spoken,
  author    = {Chao, Yuen Ren},
  title     = {A Grammar of Spoken Chinese},
  year      = {1968},
  publisher = {University of California Press},
  address   = {Berkeley and Los Angeles}
}

@book{LiThompson1981,
  author    = {Li, Charles N. and Thompson, Sandra A.},
  title     = {{Mandarin Chinese}: A Functional Reference Grammar},
  year      = {1981},
  publisher = {University of California Press},
  address   = {Berkeley},
  doi       = {10.1525/9780520352858}
}

@book{ZhuDexi1982,
  author    = {Zhu, Dexi},
  title     = {语法讲义 [Lectures on grammar]},
  year      = {1982},
  address   = {Beijing},
  publisher = {The Commercial Press}
}

@book{hua2003reduplication,
  author    = {Hua, Yuming},
  title     = {汉语重叠研究 [A study of reduplication in Chinese]},
  year      = {2003},
  address   = {Changsha},
  publisher = {Hunan People's Publishing House}
}

@article{Wang2023,
  author  = {Wang, Chen},
  title   = {A syntactic derivation of the reduplication patterns and their interpretation in {Mandarin}},
  journal = {Natural Language \& Linguistic Theory},
  volume  = {41},
  number  = {4},
  pages   = {847--877},
  year    = {2023},
  doi     = {10.1007/s11049-022-09549-y}
}

@article{lu2026deliminative,
  author  = {Lu, Yanru and M{\"u}ller, Stefan},
  title   = {Deliminative verbal reduplication in {Mandarin Chinese}},
  journal = {Journal of Linguistics},
  year    = {2026},
  pages   = {1--38},
  doi     = {10.1017/S0022226725101047},
  note    = {First published online}
}

@article{MarelliBaroni2015,
  author  = {Marelli, Marco and Baroni, Marco},
  title   = {Affixation in semantic space: Modeling morpheme meanings
             with compositional distributional semantics},
  journal = {Psychological Review},
  year    = {2015},
  volume  = {122},
  number  = {3},
  pages   = {485--515},
  doi     = {10.1037/a0039267}
}

@article{PerekHilpert2017,
  author  = {Perek, Florent and Hilpert, Martin},
  title   = {A distributional semantic approach to the periodization
             of change in the productivity of constructions},
  journal = {International Journal of Corpus Linguistics},
  year    = {2017},
  volume  = {22},
  number  = {4},
  pages   = {490--520},
  doi     = {10.1075/ijcl.16128.per}
}

@article{YangBaayen2026,
  author  = {Yang, Yi and Baayen, R. Harald},
  title   = {A quantitative study of measure words in
             {Mandarin Chinese}},
  journal = {Corpus Linguistics and Linguistic Theory},
  year    = {2026},
  note    = {Advance online publication},
  doi     = {10.1515/cllt-2025-0014}
}

@article{NikolaevEtAl2022,
  author  = {Nikolaev, Alexandre and Chuang, Yu-Ying and Baayen, R. Harald},
  title   = {A generating model for Finnish nominal inflection using distributional semantics},
  journal = {The Mental Lexicon},
  year    = {2022},
  volume  = {17},
  number  = {3},
  pages   = {368--394},
  doi     = {10.1075/ml.22008.nik}
}

@article{StupakBaayen2022,
  author  = {Stupak, Inna V. and Baayen, R. Harald},
  title   = {An inquiry into the semantic transparency and productivity of German particle verbs and derivational affixation},
  journal = {The Mental Lexicon},
  year    = {2022},
  volume  = {17},
  number  = {3},
  pages   = {422--457},
  doi     = {10.1075/ml.22012.stu}
}

@article{ShafaeiBajestanEtAl2024,
  author  = {Shafaei-Bajestan, Elnaz and Moradipour-Tari, Masoumeh and Uhrig, Peter and Baayen, R. Harald},
  title   = {The pluralization palette: Unveiling semantic clusters in English nominal pluralization through distributional semantics},
  journal = {Morphology},
  year    = {2024},
  volume  = {34},
  pages   = {369--413},
  doi     = {10.1007/s11525-024-09428-9}
}

@article{ShenBaayen2022a,
  author  = {Shen, Tian and Baayen, R. Harald},
  title   = {Adjective--noun compounds in {Mandarin}: A study on productivity},
  journal = {Corpus Linguistics and Linguistic Theory},
  year    = {2022},
  volume  = {18},
  number  = {3},
  pages   = {543--572},
  doi     = {10.1515/cllt-2020-0059}
}

@article{ShenBaayen2022b,
  author  = {Shen, Tian and Baayen, R. Harald},
  title   = {Productivity and semantic transparency:
             An exploration of word formation in {Mandarin Chinese}},
  journal = {The Mental Lexicon},
  year    = {2022},
  volume  = {17},
  number  = {3},
  pages   = {458--479},
  doi     = {10.1075/ml.22009.she}
}

@article{YangBaayen2025,
  author  = {Yang, Yi and Baayen, R. Harald},
  year    = {2025},
  title   = {Comparing the semantic structures of the lexicons of {Mandarin} and {English}},
  journal = {Language and Cognition},
  volume  = {17},
  pages   = {e10},
  doi     = {10.1017/langcog.2024.47}
}

@article{Zhang2015,
  author  = {Zhang, Niina Ning},
  title   = {The morphological expression of plurality and pluractionality in Mandarin},
  journal = {Lingua},
  volume  = {165},
  pages   = {1--27},
  year    = {2015},
  doi     = {10.1016/j.lingua.2015.07.001}
}

@article{ZhuJingsong2003,
  author  = {Zhu, Jingsong},
  title   = {形容词重叠式的语法意义 [The grammatical meaning of adjectival reduplication]},
  journal = {语文研究 [Linguistic Research]},
  year    = {2003},
  number  = {3},
  pages   = {9--17}
}

@incollection{MelloniBasciano2018,
  author    = {Melloni, Chiara and Basciano, Bianca},
  title     = {Reduplication across boundaries: The case of {Mandarin}},
  editor    = {Bonami, Olivier and Boy{\'e}, Gilles and Dal, Georgette
               and Giraudo, H{\'e}l{\`e}ne and Namer, Fiammetta},
  booktitle = {The lexeme in descriptive and theoretical morphology},
  year      = {2018},
  pages     = {325--363},
  publisher = {Language Science Press},
  address   = {Berlin},
  doi       = {10.5281/zenodo.1407013}
}

@article{zhan2019ccl,
  author  = {Zhan, Weidong and Guo, Rui and Chang, Baobao and Chen, Yirong and Chen, Long},
  title   = {The building of the {CCL} Corpus: Its design and implementation},
  journal = {Corpus Linguistics},
  year    = {2019},
  volume  = {6},
  number  = {1},
  pages   = {71--86}
}

@inproceedings{song2018directional,
  author    = {Song, Yan and Shi, Shuming and Li, Jing and Zhang, Haisong},
  title     = {Directional skip-gram: Explicitly distinguishing left and right context for word embeddings},
  booktitle = {Proceedings of the 2018 Conference of the North American Chapter of the Association for Computational Linguistics: Human Language Technologies, Volume 2 (Short Papers)},
  pages     = {175--180},
  address   = {New Orleans, Louisiana},
  publisher = {Association for Computational Linguistics},
  year      = {2018}
}

@inproceedings{mikolov2013linguistic,
  author    = {Mikolov, Tomas and Yih, Wen-tau and Zweig, Geoffrey},
  title     = {Linguistic regularities in continuous space word representations},
  booktitle = {Proceedings of the 2013 Conference of the North American Chapter of the Association for Computational Linguistics: Human Language Technologies},
  editor    = {Vanderwende, Lucy and Daum{\'e} III, Hal and Kirchhoff, Katrin},
  pages     = {746--751},
  year      = {2013},
  address   = {Stroudsburg, PA},
  publisher = {Association for Computational Linguistics},
  url       = {https://aclanthology.org/N13-1090}
}

@article{mikolov2013distributed,
  title={Distributed representations of words and phrases and their
compositionality},
  author={Mikolov, Tomas and Sutskever, Ilya and Chen, Kai and
Corrado, Greg S and Dean, Jeff},
  journal={Advances in neural information processing systems},
  volume={26},
  year={2013}
}

@inproceedings{macqueen1967some,
  title={Some methods of classification and analysis of multivariate
observations},
  author={MacQueen, James B},
  booktitle={Proc. of 5th berkeley symposium on math. stat. and prob.},
  pages={281--297},
  year={1967}
}

@book{VenablesRipley2002,
  author    = {Venables, William N. and Ripley, Brian D.},
  title     = {Modern Applied Statistics with S},
  edition   = {4},
  year      = {2002},
  publisher = {Springer},
  address   = {New York}
}

@article{Maatentsne,
  author  = {van der Maaten, Laurens and Hinton, Geoffrey},
  year    = {2008},
  title   = {Visualizing data using {t-SNE}},
  journal = {Journal of Machine Learning Research},
  volume  = {9},
  pages   = {2579--2605}
}

@article{westbury2019wordcategory,
  author  = {Westbury, Chris and Hollis, Geoff},
  title   = {Conceptualizing syntactic categories as semantic categories: Unifying part-of-speech identification and semantics using co-occurrence vector averaging},
  journal = {Behavior Research Methods},
  year    = {2019},
  volume  = {51},
  pages   = {1371--1398},
  doi     = {10.3758/s13428-018-1118-4}
}

@article{westbury2015emotion,
  author  = {Westbury, Chris and Keith, Jeff and Briesemeister, Benny B. and Hofmann, Markus J. and Jacobs, Arthur M.},
  title   = {Avoid violence, rioting, and outrage; approach celebration, delight, and strength: Using large text corpora to compute valence, arousal, and the basic emotions},
  journal = {The Quarterly Journal of Experimental Psychology},
  year    = {2015},
  volume  = {68},
  number  = {8},
  pages   = {1599--1622},
  doi     = {10.1080/17470218.2014.970204}
}

@article{westburywurm2022self,
  author  = {Westbury, Chris and Wurm, Lee H.},
  title   = {Is it you you're looking for? Personal relevance as a principal component of semantics},
  journal = {The Mental Lexicon},
  year    = {2022},
  volume  = {17},
  number  = {1},
  pages   = {1--33},
  doi     = {10.1075/ml.20031.wes}
}

@article{xu2008affective,
  author   = {Xu, Linhong and Lin, Hongfei and Pan, Yu and Ren, Hui and Chen, Jianmei},
  title    = {Qinggan cihui benti de gouzao [Constructing the Affective Lexicon Ontology]},
  journal  = {Qingbao Xuebao [Journal of the China Society for Scientific and Technical Information]},
  year     = {2008},
  volume   = {27},
  number   = {2},
  pages    = {180--185},
  doi      = {10.3969/j.issn.1000-0135.2008.02.004},
  language = {Chinese}
}

@book{XiaoRaysonMcEnery2009,
  author    = {Xiao, Richard and Rayson, Paul and McEnery, Tony},
  title     = {A Frequency Dictionary of Mandarin Chinese: Core Vocabulary for Learners},
  publisher = {Routledge},
  address   = {London},
  year      = {2009}
}

@article{Ekman1992,
  author  = {Ekman, Paul},
  title   = {An argument for basic emotions},
  journal = {Cognition and Emotion},
  year    = {1992},
  volume  = {6},
  number  = {3-4},
  pages   = {169--200},
  doi     = {10.1080/02699939208411068}
}

@article{gaver1993sound,
  author  = {Gaver, William W.},
  title   = {What in the world do we hear? An ecological approach to auditory event perception},
  journal = {Ecological Psychology},
  volume  = {5},
  number  = {1},
  pages   = {1--29},
  year    = {1993},
  doi     = {10.1207/s15326969eco0501_1}
}

@inproceedings{drozd2016word,
  author    = {Drozd, Aleksandr and Gladkova, Anna and Matsuoka, Satoshi},
  title     = {Word embeddings, analogies, and machine learning: Beyond king - man + woman = queen},
  booktitle = {Proceedings of {COLING} 2016, the 26th International Conference on Computational Linguistics: Technical Papers},
  editor    = {Matsumoto, Yuji and Prasad, Rashmi},
  pages     = {3519--3530},
  year      = {2016},
  publisher = {The {COLING} 2016 Organizing Committee},
  url       = {https://aclanthology.org/C16-1332}
}

@article{boleda2020distributional,
  author  = {Boleda, Gemma},
  title   = {Distributional semantics and linguistic theory},
  journal = {Annual Review of Linguistics},
  year    = {2020},
  volume  = {6},
  pages   = {213--234},
  doi     = {10.1146/annurev-linguistics-011619-030303},
  eprint  = {1905.01896},
  archivePrefix = {arXiv}
}

@article{ShafaeiBajestanUhrigBaayen2022,
  author  = {Shafaei-Bajestan, Elnaz and Uhrig, Peter and Baayen, R. Harald},
  title   = {Making sense of spoken plurals},
  journal = {The Mental Lexicon},
  year    = {2022},
  volume  = {17},
  number  = {3},
  pages   = {337--367},
  doi     = {10.1075/ml.22011.sha}
}

@article{ChuangEtAl2022,
  author  = {Chuang, Yu-Ying and Brown, Dunstan and Baayen, R. Harald and Evans, Roger},
  title   = {Paradigm gaps are associated with weird ``distributional semantics'' properties: Russian defective nouns and their case and number paradigms},
  journal = {The Mental Lexicon},
  year    = {2022},
  volume  = {17},
  number  = {3},
  pages   = {395--421},
  doi     = {10.1075/ml.22013.chu}
}

@article{Mohiuddin2020,
  author  = {Mohiuddin, Tasnim and Joty, Shafiq},
  year    = {2020},
  title   = {Unsupervised word translation with adversarial autoencoder},
  journal = {Computational Linguistics},
  volume  = {46},
  number  = {2},
  pages   = {257--288},
  doi     = {10.1162/coli_a_00374}
}

@article{JorgeBotana2018,
  author  = {Jorge-Botana, Guillermo and Olmos, Ricardo and Luz{\'o}n, Jos{\'e} M.},
  year    = {2018},
  title   = {Word maturity indices with latent semantic analysis: Why, when, and where is Procrustes rotation applied?},
  journal = {WIREs Cognitive Science},
  volume  = {9},
  number  = {1},
  pages   = {e1457},
  doi     = {10.1002/wcs.1457}
}

@article{Kuhn1955,
  author  = {Kuhn, Harold W.},
  year    = {1955},
  title   = {The Hungarian method for the assignment problem},
  journal = {Naval Research Logistics Quarterly},
  volume  = {2},
  number  = {1--2},
  pages   = {83--97},
  doi     = {10.1002/nav.3800020109}
}

@manual{Oksanen2022,
  title  = {vegan: Community Ecology Package},
  author = {Oksanen, Jari and Simpson, Gavin L. and Blanchet, F. Guillaume and Kindt, Roeland and Legendre, Pierre and Minchin, Peter R. and O'Hara, R. B. and Solymos, Peter and Stevens, M. Henry H. and Szoecs, Eduard and Wagner, Helene and Barbour, Michael and Bedward, Michael and Bolker, Benjamin and Borcard, Daniel and Carvalho, Gustavo and Chirico, Michael and De Caceres, Miquel and Durand, Sebastien and Evangelista, Heloisa B. A. and FitzJohn, Rich and Friendly, Michael and Furneaux, Brendan and Hannigan, Geoffrey and Hill, Mark O. and Lahti, Leo and McGlinn, Daniel and Ouellette, Marie-H{\'e}l{\`e}ne and Ribeiro Cunha, Eduardo and Smith, Tyler and Stier, Adrian and Ter Braak, Cajo J. F. and Weedon, James},
  year   = {2022},
  note   = {R package version 2.6-4}
}

@article{ZhuJingsong1998,
  author  = {Zhu, Jingsong},
  title   = {动词重叠式的语法意义 [The grammatical meaning of verbal reduplication]},
  journal = {中国语文 [Studies of the Chinese Language]},
  year    = {1998},
  number  = {5},
  pages   = {378--386}
}

\end{document}